\documentclass[letterpaper, 10 pt, journal, twoside]{IEEEtran}

\usepackage{graphics} 
\usepackage{epsfig} 
\usepackage{times} 
\usepackage{amsmath} 
\usepackage{amssymb}  
\usepackage[hang]{subfigure} 
\usepackage{cite}
\usepackage{siunitx}
\usepackage{soul}
\usepackage{color}
\usepackage{xcolor}

\soulregister\ref7  
\soulregister\cite7 
\soulregister\eqref7
\soulregister\prettyref7

\usepackage{multirow}
\usepackage{makecell}
\usepackage{booktabs}
\usepackage{threeparttable}
\usepackage[linesnumbered,ruled,vlined]{algorithm2e}

\bstctlcite{IEEEexample:BSTcontrol}

\usepackage{prettyref}
\newrefformat{fig}{Fig.~\ref{#1}}
\newrefformat{sec}{Sec.~\ref{#1}}
\newrefformat{tab}{Tab.~\ref{#1}}
\newrefformat{algo}{Algo.~\ref{#1}}
\newrefformat{eq}{(\ref{#1})}
\newrefformat{alli}{\emph{line}~\ref{#1}}

\begin{document}
\title{\LARGE \bf
A Generalized Continuous Collision Detection Framework of Polynomial Trajectory for Mobile Robots in Cluttered Environments
}
%
\author{Zeqing Zhang$^{1,2}$, Yinqiang Zhang$^{1}$, Ruihua Han$^{1}$, Liangjun Zhang$^{2}$ and Jia Pan$^{1,\dagger}$%
\thanks{$^{1}$ Department of Computer Science, The University of Hong Kong, Hong Kong, China.}
\thanks{$^{2}$ Robotics and Autonomous Driving Lab of Baidu Research, Beijing, China.}
\thanks{$\dagger$ Corresponding author}
\thanks{Email: \tt\small \{zzqing, zyq507, hanrh\}@connect.hku.hk, liangjunzhang@baidu.com, jpan@cs.hku.hk}

}

\markboth{IEEE ROBOTICS AND AUTOMATION LETTERS. PREPRINT VERSION. ACCEPTED JUNE, 2022}
{Zhang \MakeLowercase{\textit{et al.}}: A Generalized Continuous Collision Detection Framework}

\maketitle
\thispagestyle{empty}
\pagestyle{empty}

\begin{abstract}
In this paper, we introduce a generalized continuous collision detection (CCD) framework for the mobile robot along the polynomial trajectory in cluttered environments including various static obstacle models. Specifically, we find that the collision conditions between robots and obstacles could be transformed into a set of polynomial inequalities, whose roots can be efficiently solved by the proposed solver. In addition, we test different types of mobile robots with various kinematic and dynamic constraints in our generalized CCD framework and validate that it allows the provable collision checking and can compute the exact time of impact. Furthermore, we combine our architecture with the path planner in the navigation system. Benefiting from our CCD method, the mobile robot is able to work safely in some challenging scenarios.
\end{abstract}

\begin{IEEEkeywords}
Collision avoidance, motion and path planning, robot safety.
\end{IEEEkeywords}

\section{Introduction}
\label{sec:intro}
\IEEEPARstart{C}{urrently} a large number of mobile robots have been used in industrial applications to reduce the labor costs and improve the productivity.
%
Since mobile robots basically work in the shared workspace, the \emph{collision avoidance} with each other and the environment becomes a significant issue for the planner when coordinating the swarm. 
For ground mobile robots, such as the automated guided vehicle (AGV), commercial planners normally employ different scheduling schemes to avoid collisions at some potential conflict points, such as the crossroad of planned paths \cite{zhang2021efficient}. Some decentralized planners take advantage of the reinforcement learning to ensure safety in the multi-agent system \cite{han2022reinforcement}.
For aerial mobile robots, such as the quadrotor, the trajectory planner can generate the collision free trajectories in the pre-built Euclidean signed distance field \cite{ratliff2009chomp} or the flight corridor \cite{liu2017planning}.

Among above collision avoidance strategies, there is a key problem to be well solved, that is the \emph{collision detection} to determine the time of impact (ToI) of the moving robot against other objects.
A conventional framework of that is \emph{discretized}, where the trajectory of robot would be discretized into a set of time instants, and then some collision checking methods, such as separating axis theorem and Gilbert–Johnson–Keerthi algorithm (GJK)~\cite{ref_ericson2004real}, would be used to test conflicts at each sampled time. Since this method may miss the collision between the sampled instants, resulting in the known tunneling phenomenon \cite{ref_ericson2004real}, the \emph{continuous collision detection} (CCD) is required.

Although there are several methods implementing the CCD in computer graphics and games \cite{wang2021large}, the CCD for robotics, especially the mobile robot, remains the following three challenges. 
\begin{figure}[!tb]
    \centering
    \includegraphics[width=0.4\textwidth]{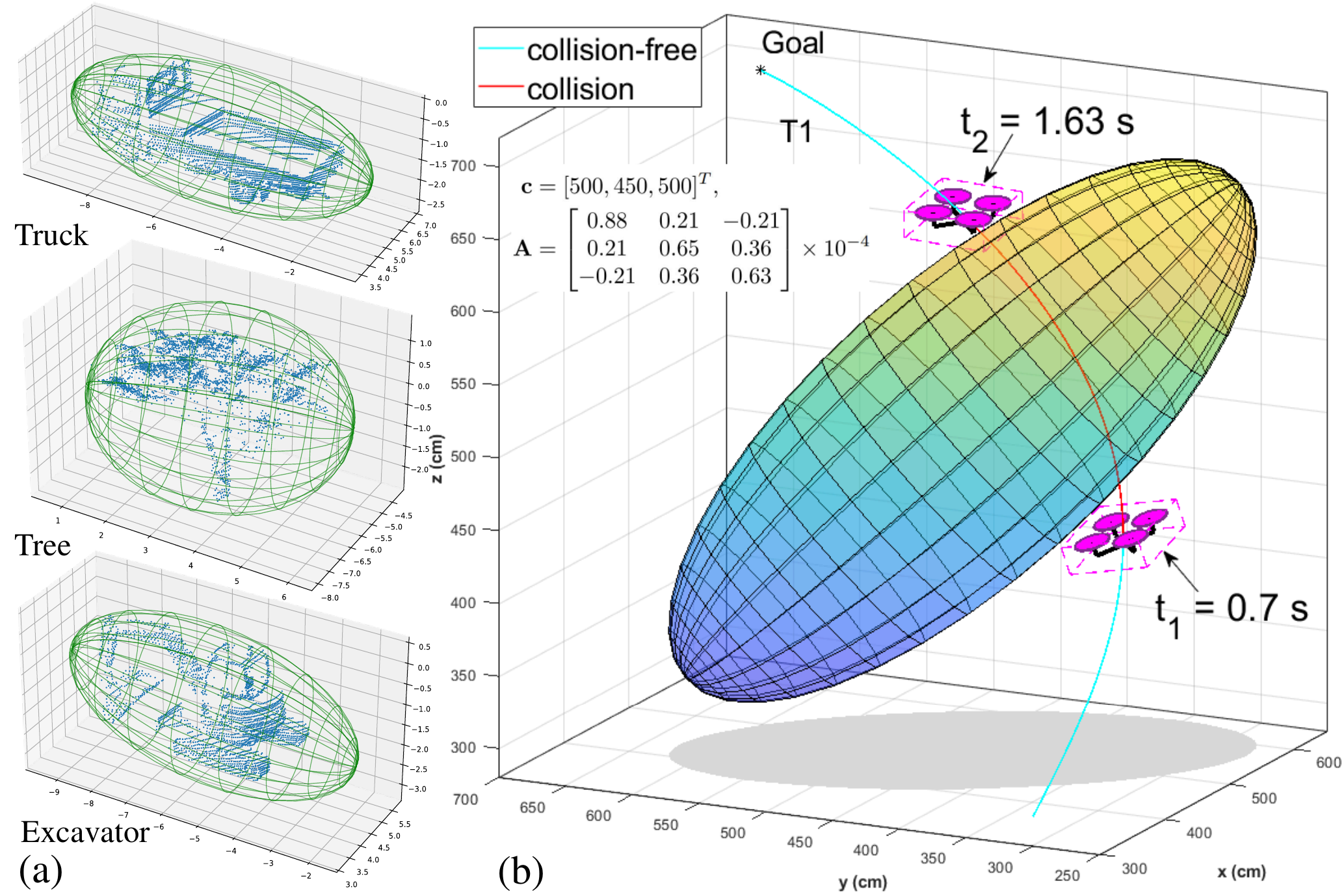}
    \vspace{-10pt}
    \caption{(a) The point clouds of some complicated obstacles from \cite{de2013unsupervised} can be enclosed using the outer L{\"o}wner-John ellipsoid \cite{ben2001lectures}. (b) In our CCD, the collision interval between a box-shaped quadrotor along the trajectory T1 against the ellipsoidal obstacle is determined by $t \in [0.6959, 1.6347]$, as shown in red curve. The polynomial trajectory is given in \prettyref{eq:uavTraj1}.}
    \label{fig:uav_elip}
    \vspace{-10pt}
\end{figure}

First, robots need to be enclosed abstractly by simple geometric shapes, called \textbf{bounding objects}. The simplest way is using a circle or sphere to cover the mobile robot, since collision tests between moving circles/spheres are trivial and can be operated in a fast and effective manner. To be compact, few methods would employ the ellipse/ellipsoid to model the robots \cite{choi2006continuous}, largely because it is not easy and efficient to do the interference checking. However, in some cases, such as the automated warehouse, it needs to tightly envelop mobile robots to save operation space in the limited working area. Thus the bounding box becomes a better choice, whereas leading to difficulties about CCD for the bounding box. In addition, for some mobile robots, it is not suitable to do conflict detection using bounding boxes, such as the collision checking for the cable-driven parallel robot (CDPR) \cite{bury2019continuous}, because its potential interference mainly comes from straight cables with the environment.

Secondly, the \textbf{environment representation} is still an open problem. It is straightforward to represent the surroundings by the occupancy grid map \cite{elfes1989using}, whose dense elements are featured by the  probability of occupancy. The maps defined by point-clouds are other sparse representations. However, these discretized information can not be used directly in the CCD.

Thirdly, unlike the CCD in computer graphics, where the movement of objects can be defined easily and separately, it is inevitable to consider \textbf{robotic constraints} from nonlinear kinematics and dynamics for CCD of mobile robots in practice. For instance, the nonholonomic AGV can not change direction arbitrarily, and the motion of quadrotor needs to comply with its dynamic conditions as well. These challenges provide difficulties for the CCD of mobile robots.

\begin{figure}[!tb]
    \centering
    \vspace{-5pt}
    \includegraphics[width=0.45\textwidth]{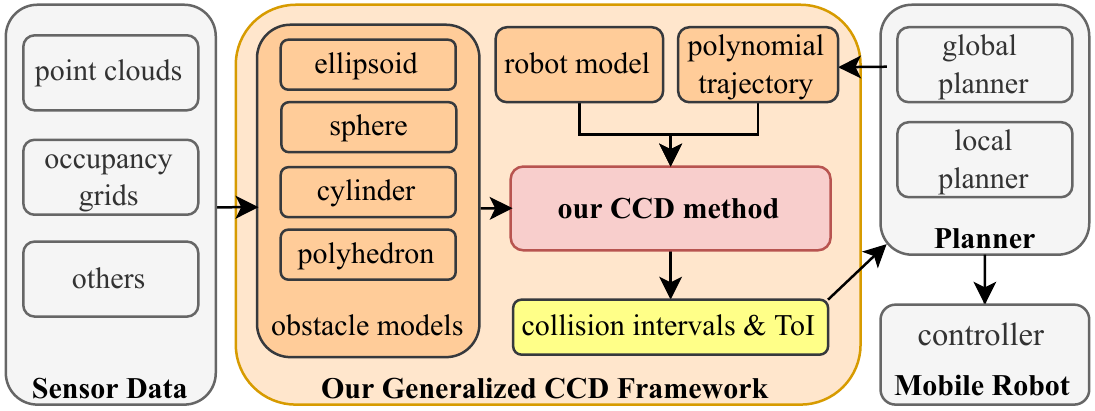}
    \vspace{-5pt}
    \caption{The workflow of our generalized CCD framework, whose role in the real navigation system for mobile robots is presented as well.}
    \label{fig:workflow}
    \vspace{-10pt}
\end{figure}

To this end, we propose a generalized CCD framework for mobile robots along polynomial trajectories in the cluttered environment, composed of proposed obstacle models. The entire workflow of the framework is given in \prettyref{fig:workflow}. We investigate the collision cases of the edge and find collision conditions between robot's edges and obstacles can be transformed into a set of polynomial inequalities, whose roots provide the exact collision instants for moving robots in both translation and orientation.
In this paper, the translational trajectories are specified by polynomials, which is reasonable since planners normally generate the analytical trajectories in the polynomial or piece-wise polynomial forms. Meanwhile, the robot's orientation can be determined according to its nonlinear kinematic or dynamic constraints. In addition, our method also works for non-polynomial motion by its Taylor series within the acceptable approximation error. 
Finally, we employ three different types of mobile robots, i.e., CDPR, quadrotor and AGV, to validate our generalized CCD framework, and further combine it with the wildly used local planner, dynamic window approach (DWA) \cite{dwa}, as an example to explore its application potential.

\noindent{\textbf{Main contributions}}: 
\begin{itemize}
    \item We propose an efficient CCD method to compute the ToI and collision intervals by transforming collision conditions about edges into polynomial inequalities, whose roots can be efficiently solved by the proposed algorithm.
    \item We employ a family of obstacle models, including the ellipsoid, sphere, cylinder and polyhedron, which could be extracted from the perceived sensor data of environments. 
    \item We respect the nonlinear kinematic and dynamic constraints of mobile robots, and successfully combine our method into a path planner to explore further challenging scenarios. 
\end{itemize}

The rest of the paper is organized as follows. \prettyref{sec:related} reviews the related work about CCD and \prettyref{sec:bkg} describes the mobile robot model and collision cases about the edge. \prettyref{sec:collisionChecking} reformulates the collision conditions between robot and obstacle models into a set of polynomial inequalities. Then a general coefficient determination approach and an efficient polynomial roots-finding algorithm are given in \prettyref{sec:application}. Extensive simulation experiments are given in \prettyref{sec:experiment}, and \prettyref{sec:conclusion} concludes this paper.

\section{Related work}
\label{sec:related}
To determine the collision, the objects should be first enclosed by some simple geometric primitives, such as circle/sphere, axis-aligned/oriented bounding boxes, to which the crude collision detection methods can be applied. Then if simple geometries overlap, some advanced collision detection techniques would be used for objects with more complicated shapes. Since we investigate the basic collision of edges, the mobile robot will be modeled by a bunch of edges in our framework. In addition, we can represent the obstacles by several shapes according to the accuracy requirements of applications.

To determine the ToI and avoid the tunneling, various CCD methods have been implemented.
A simple brute-force solution is increasing the sample rate, i.e., \emph{supersampling}, which not only causes time intensive but still has the chance to miss the collision.
The \emph{binary search} is another simple yet effective technique using the bisection method to narrow down the possible collision time interval until it has an acceptable error. However, this method does not solve the tunneling issue, since no conflict occurs both at the successive instants before and after the collision. \emph{Ray casting} \cite{van2003collision} is typically used in many shooting games to check the impact of bullet. Nevertheless, it is mainly good at the linear motion that can be presented by rays. In comparison, our method can tackle CCD for robots with polynomial trajectories in translation and orientation simultaneously, complying with the robotic kinematic and dynamic constraints as well.

In addition, if the shape of objects is simple, such as the sphere or triangle, it is possible to find the \emph{analytical solution} of ToI between pair of moving objects \cite{glassner2013graphics}. For general cases, collisions can be checked between the \emph{swept volume} of an object with a specific custom bounding volume hierarchy \cite{herman1986fast}. Furthermore, \emph{conservative advancement} \cite{pan2012collision} is another known approach. 
However, it often gives too conservative results when objects are close together, but not touching. For our method, it provides another analytical solution to CCD by transforming the collision conditions into a set of polynomial inequalities, whose roots can be efficiently solved by the proposed roots-finding algorithm. Also our CCD can guarantee to find every collision and calculate the exact ToI without iterations.

\section{Background}
\label{sec:bkg}
\subsection{Geometric Model of Mobile Robots}
\begin{figure}
    \centering
    \subfigure[Mobile robot model]{
    \label{fig:uav3d}
    \includegraphics[width=0.17\textwidth]{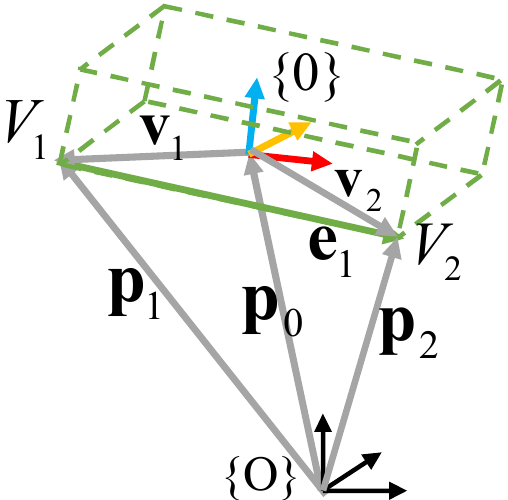}}~\hfil
    \subfigure[Edge-point]{
    \label{fig:seg_pt}
    \includegraphics[width=0.2\textwidth]{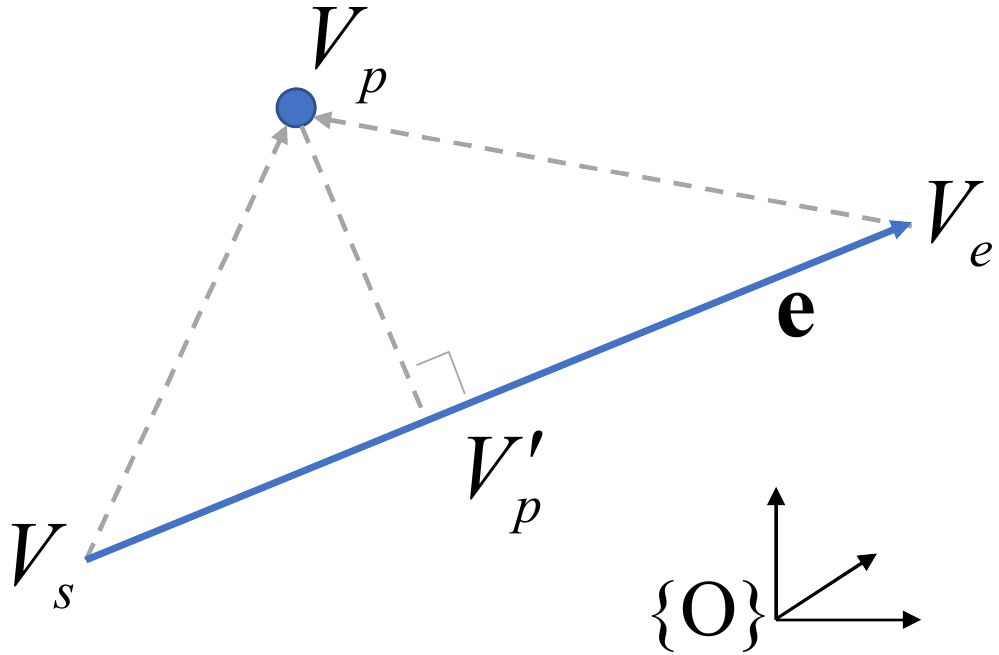}}~\vfil
	\subfigure[Edge-edge]{
    \label{fig:seg_seg}
    \includegraphics[width=0.2\textwidth]{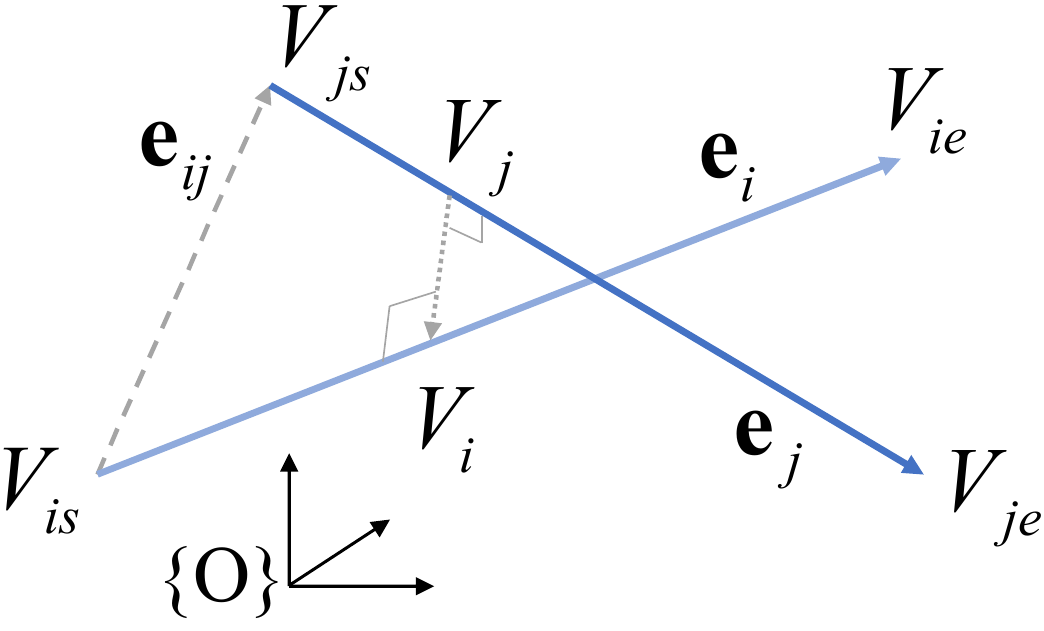}}~\hfil
    \subfigure[Edge-triangular surface]{
    \label{fig:seg_tri}
    \includegraphics[width=0.23\textwidth]{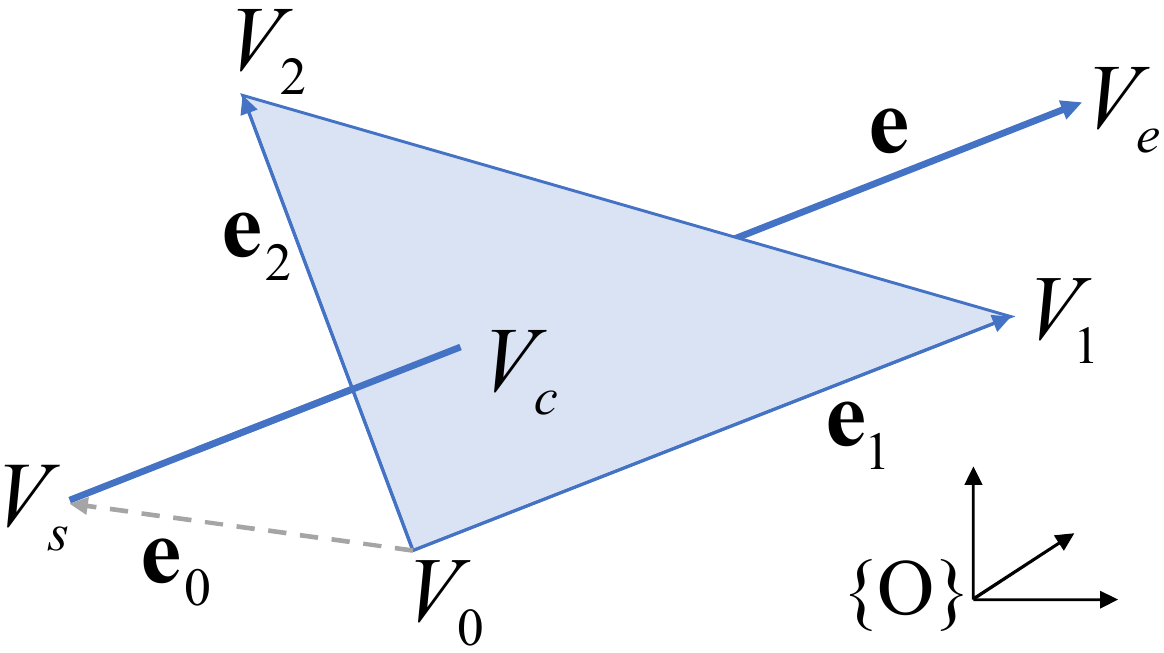}}
    \vspace{-5pt}
    \caption{Robot model and collision cases.}
    \vspace{-10pt}
\end{figure}

In this paper, the mobile robot is modeled as a rigid body with $\varepsilon$ edges (in green), as presented in \prettyref{fig:uav3d}.
Here, the edge is defined as the line segment with finite length and all vectors below are assumed to be column vectors.

Based on the geometric relationship in \prettyref{fig:uav3d}, the position of vertex $V_1$ (i.e., $\mathbf{p}_1 := \overrightarrow{OV_1}$) and edge vector $\overrightarrow{V_1V_2}$ (i.e., $\mathbf{e}_1 :=  \overrightarrow{OV_2} - \overrightarrow{OV_1}$) can be formulated in the world frame $\{O\}$ as
\begin{align}
    \mathbf{p}_1 &= \mathbf{p}_{0} + R~\mathbf{v}_1,
	\label{eq:vertex2D}\\
    \mathbf{e}_1 &= \mathbf{p}_2 - \mathbf{p}_1 = R~(\mathbf{v}_2 - \mathbf{v}_1), \label{eq:edge2D} 
\end{align}
where $\mathbf{p}_{0} = [x,y,z]^T$ refers to the position of robot in the world frame $\{O\}$, and $R$ is the rotation matrix from the body frame $\{0\}$ attached on the model to the world frame $\{O\}$. In addition, $\mathbf{v}_1$ and $\mathbf{v}_2$ are position vectors of vertices $V_1$ and $V_2$ expressed in the body frame $\{0\}$, which are constant and determined by the model geometry. Here we employ the ZYX Euler angles to formulate its rotation matrix:
\begin{align}\label{eq:rotMat3d}
    R = 
    \begin{bmatrix}
		&c\theta c\psi &s\phi s\theta c\psi - c\phi s\psi &c\phi s\theta c\psi + s\phi s\psi\\
		&c\theta s\psi &s\phi s\theta s\psi + c\phi c\psi &c\phi s\theta s\psi - s\phi c\psi\\
		&-s\theta &s\phi c\theta &c\phi c\theta
	\end{bmatrix},
\end{align}
where $\phi\in[-\pi,\pi]$ for roll, $\theta\in[-\pi/2,\pi/2]$ for pitch, $\psi\in[-\pi,\pi]$ for yaw. And $s$ and $c$ stand for $\sin$ and $\cos$, respectively.
\vspace{-10pt}

\subsection{Collision Cases}\label{sec:sub_collicases}
As shown in \prettyref{fig:seg_pt}-\subref{fig:seg_tri}, we will consider $3$ different collision cases about the edge of robot model against the point, another edge and the triangular surface, respectively. The safe distance between robot and obstacles can be prescribed as $\underline{d}$.
\subsubsection{Edge-Point}\label{sec:subsub_EP}
As in \prettyref{fig:seg_pt}, the point $V_p'$ indicates the projection of the point $V_p$ onto the line contacting the edge $\mathbf{e}:=\overrightarrow{V_sV_e}$. Based on the relative position of $V_p'$ with respect to (w.r.t.) the edge $\mathbf{e}$, then the \textbf{d}istance between the \textbf{E}dge $\mathbf{e}$ and the \textbf{P}oint $V_p$ can be calculated by the following function: 
\begin{align}\label{eq:edge_pt}
    &\mathbf{dEP}(\mathbf{e},V_p) = \nonumber\\
    &\begin{cases}
        \|\overrightarrow{V_sV_p}\|, &~\text{if}~ \overrightarrow{V_sV_p} \cdot \mathbf{e} \leqslant 0  \\
        {\|\mathbf{e} \times \overrightarrow{V_sV_p}\|}/{\|\mathbf{e}\|}, &~\text{if}~ 0 < \overrightarrow{V_sV_p} \cdot \mathbf{e} < \mathbf{e} \cdot \mathbf{e} \\
        \|\overrightarrow{V_eV_p}\|, &~\text{if}~ \mathbf{e} \cdot \mathbf{e} \leqslant \overrightarrow{V_sV_p} \cdot\mathbf{e}
    \end{cases}
\end{align}
where $\cdot$ is the dot product. See Sec. 5.1.2 in \cite{ref_ericson2004real} for more details. As such, the collision occurs when $\mathbf{dEP}(\mathbf{e},V_p) < \underline{d}$.

\subsubsection{Edge-Edge}\label{sec:sub_edgeedge}
Since we will check collision for each pair of edges, then if certain pair of edges $\mathbf{e}_i$ and $\mathbf{e}_j$ are in parallel, we would not consider the potential interference just between this pair of edges. 

For two non-parallel edges (i.e., $\mathbf{e}_i \times \mathbf{e}_j \neq \mathbf{0}$), as shown in \prettyref{fig:seg_seg}, $V_i$ and $V_j$ constitute the common perpendicular segment. 
According to the geometric relationship, it has
\begin{align} \label{eq:edge_edge}
    \overrightarrow{V_{is}V_i} = \overrightarrow{V_{is}V_{js}} + \overrightarrow{V_{js}V_j} + \overrightarrow{V_jV_i}.
\end{align}
As such, if we define $ \mathbf{e}_{ij} := \overrightarrow{V_{is}V_{js}}$, $ t_i \mathbf{e}_i := \overrightarrow{V_{is}V_i}$, $ t_j \mathbf{e}_j := \overrightarrow{V_{js}V_j}$ and $ t_p(\mathbf{e}_i \times \mathbf{e}_j) := \overrightarrow{V_jV_i}$, where $t_i, t_j, t_p$ are scalar parameters, then \prettyref{eq:edge_edge} can be rewritten as
\begin{align}\label{eq:compact_edge_edge}
    \mathbf{M}\mathbf{t} = \mathbf{e}_{ij},
\end{align}
where $\mathbf{M} = [\mathbf{e}_i,\ -\mathbf{e}_j,\ -(\mathbf{e}_i \times \mathbf{e}_j)]$ and $\mathbf{t} = [t_i,\ t_j,\ t_p]^T$. Since $ (\mathbf{e}_i \times \mathbf{e}_j) \perp \mathbf{e}_i, (\mathbf{e}_i \times \mathbf{e}_j) \perp \mathbf{e}_j$ and $\mathbf{e}_i \times \mathbf{e}_j \neq \mathbf{0}$ by definition, then \eqref{eq:compact_edge_edge} can be uniquely solved by
\begin{align}\label{eq:4itemsEE}
    \mathbf{t} &= \mathbf{M}^{-1} \mathbf{e}_{ij} \nonumber\\
	           &= \frac{1}{\det{\mathbf{M}}}
				\begin{bmatrix}
				\det{[\mathbf{e}_{ij}, -\mathbf{e}_j, -\mathbf{e}_i \times \mathbf{e}_j]}\\
				\det{[\mathbf{e}_i, \mathbf{e}_{ij}, -\mathbf{e}_i \times \mathbf{e}_j]}\\
				\det{[\mathbf{e}_i, -\mathbf{e}_j, \mathbf{e}_{ij}]}\\
				\end{bmatrix}
				:= \frac{1}{u_0}
				\begin{bmatrix}
					u_1\\ u_2 \\ u_3
				\end{bmatrix},
\end{align}
where $\det$ refers to the determinant. 

Based on the relative positions of $V_i$ and $V_j$ w.r.t. $\mathbf{e}_i$ and $\mathbf{e}_j$, there are $9$ cases to calculate the distance between two edges \cite{lumelsky1985fast}. For simplicity, the \textbf{d}istance between \textbf{E}dge $\mathbf{e}_i$ and \textbf{E}dge $\mathbf{e}_j$ can be determined as follows.
\begin{align}\label{eq:dEE}
    &\mathbf{dEE}(\mathbf{e}_i,\mathbf{e}_j) = \nonumber\\
    &\begin{cases}
    |t_p|\|\mathbf{e}_i \times \mathbf{e}_j\|,&~\text{if} ~0\leqslant t_i\leqslant 1, \ 0\leqslant t_j\leqslant 1 \\ 
    \mathbf{dEP}(\mathbf{e}_j,V_{is}), &~\text{if}~  t_i< 0, \ 0\leqslant t_j\leqslant 1\\
    \mathbf{dEP}(\mathbf{e}_j,V_{ie}),&~\text{if}~  1< t_i, \ 0\leqslant t_j\leqslant 1 \\ 
    \mathbf{dEP}(\mathbf{e}_i,V_{js}),&~\text{if}~ 0\leqslant t_i\leqslant 1, \ t_j< 0 \\ 
    \mathbf{dEP}(\mathbf{e}_i,V_{je}),&~\text{if}~ 0\leqslant t_i\leqslant 1, \ 1< t_j \\ 
    \|\overrightarrow{V_{is}V_{js}}\|, &~\text{if}~  t_i< 0, \ t_j< 0\\
    \|\overrightarrow{V_{ie}V_{js}}\|, &~\text{if}~  1< t_i, \ t_j< 0\\
    \|\overrightarrow{V_{is}V_{je}}\|, &~\text{if}~  t_i< 0, \ 1< t_j\\
    \|\overrightarrow{V_{ie}V_{je}}\|. &~\text{if}~  1< t_i, \ 1< t_j\\
    \end{cases}
\end{align}
Thus, the collision occurs when $\mathbf{dEE}(\mathbf{e}_i,\mathbf{e}_j) < \underline{d}$.

\subsubsection{Edge-Triangular Surface}
Defining a triangle by its vertices $V_0$, $V_1$ and $V_2$ in \prettyref{fig:seg_tri}, the intersection point $V_c$ between edge $\mathbf{e}$ and the triangle can be determined by
\begin{align}\label{eq:edge_tri}
    \overrightarrow{OV_s} + k\mathbf{e} = (1-k_1-k_2)\overrightarrow{OV_0}+k_1\overrightarrow{OV_1}+k_2\overrightarrow{OV_2}
\end{align}
where $k$ is a scalar parameter, and $k_1$, $k_2$ are barycentric coordinates of $V_c$ on the triangle \cite{rayTri}. Let $\mathbf{e}_{1} := \overrightarrow{OV_{1}}-\overrightarrow{OV_{0}}$, $\mathbf{e}_{2} := \overrightarrow{OV_{2}}-\overrightarrow{OV_{0}}$ and $\mathbf{e}_0 := \overrightarrow{OV_s}-\overrightarrow{OV_{0}}$, then \prettyref{eq:edge_tri} becomes
\begin{align}\label{eq:compact_edge_tri}
    \mathbf{Q}\mathbf{k} = \mathbf{e}_0
\end{align}
where $\mathbf{Q} = [-\mathbf{e},\ \mathbf{e}_{1},\ \mathbf{e}_{2}]$ and $\mathbf{k} = [k,~k_1,~k_2]^T$. The linear equation \eqref{eq:compact_edge_tri} is solvable if and only if $\mathbf{Q}$ is invertible, i.e., $\det{\mathbf{Q}} \neq 0$, thus
\begin{align}\label{eq:4itemsTri}
    \mathbf{k} &= \mathbf{Q}^{-1}\mathbf{e}_0 \nonumber\\
	&=\frac{1}{\det{\mathbf{Q}}}
    \begin{bmatrix}
    	\det{[\mathbf{e}_0, \mathbf{e}_{1}, \mathbf{e}_{2}]}\\
    	\det{[-\mathbf{e}, \mathbf{e}_0, \mathbf{e}_{2}]}\\
        \det{[-\mathbf{e}, \mathbf{e}_{1}, \mathbf{e}_0]}\\
    \end{bmatrix}
    := \frac{1}{v_3}
    \begin{bmatrix}
        v_0\\ v_1 \\ v_2
    \end{bmatrix}.
\end{align}
In this case, the interference conditions can be stated as:
\begin{align}\label{eq:cond_segTri}
    0\leqslant k \leqslant1,~~ 0\leqslant k_1,~~ 0\leqslant k_2,~~ k_1 + k_2 \leqslant 1.
\end{align}
Since we will check collision for each edge against the triangular surface, if $\mathbf{e}$ is parallel to the triangle (i.e., $\det{\mathbf{Q}} = 0$), no potential interference is considered in this paper.

\section{Collision Checking}
\label{sec:collisionChecking}
\subsection{Robot Edge and Trajectory as Polynomials}
\label{sec:subsec_path}
In this paper, we consider the mobile robot translates in the polynomial forms, i.e., $x(t), y(t), z(t)$ in $\mathbf{p}_{0}$ of \prettyref{eq:vertex2D} are all degree-$n$ univariate polynomial equations of time $t$:
\begin{align}\label{eq:xyzPsiPoly3D}
    x(t),y(t),z(t) \in P(t^n), t\in[t_s,t_e].
\end{align}
For orientation movement, some robots can be defined independently, e.g., CDPR, thus they can make rational motion \cite{choi2006continuous}, where each element in rotation matrix \prettyref{eq:rotMat3d} is presented by polynomials. 
But for other mobile robots, their orientations are mainly coupled with translation trajectories due to robotic kinematics and dynamics.
So, here we will estimate parameters in \prettyref{eq:rotMat3d} as a set of degree-$p$ polynomial equations w.r.t. $t$ according to robotic constraints and polynomial trajectories \prettyref{eq:xyzPsiPoly3D}, that is,
\begin{align}\label{eq:sincosFit3d}
    \begin{split}
        {\sin}\mathcal{X} \cong f_{\sin}(t) &\in P(t^p),\ 
        {\cos}\mathcal{X} \cong f_{\cos}(t) \in P(t^p),\\ &\mathcal{X} \in \{\phi(t), \theta(t), \psi(t) : \prettyref{eq:rotMat3d}\}.
    \end{split}
\end{align}
Specifically, we employ the $polyfit$ function in MATLAB to do the estimation, and the estimation error is limited within $1$ degree. More details are given in \prettyref{sec:experiment}. For brevity, we will use the identical degree $p$ for all polynomial equations in \prettyref{eq:sincosFit3d}, even if they may have different polynomial degrees.

In the following, we substitute \prettyref{eq:sincosFit3d} into \prettyref{eq:rotMat3d} and further put \prettyref{eq:rotMat3d} and \prettyref{eq:xyzPsiPoly3D} into \prettyref{eq:vertex2D} and \prettyref{eq:edge2D}, then vectors of vertex $V_1$ and edge $\mathbf{e}_1$ would be converted to following polynomial equations w.r.t. $t$:
\begin{align}\label{eq:vertex_seg_t_3d}
\begin{split}
    \mathbf{p}_1 &\in 
        [P(t^{\overline{q}}),  P(t^{\overline{q}}), P(t^{\underline{q}})]^T,\\
    \mathbf{e}_1 &\in 
        [P(t^{3p}), P(t^{3p}), P(t^{2p})]^T,\\
    \overline{q} &= \max(n, 3p),
    \underline{q} = \max(n, 2p).
\end{split}
\end{align}

\begin{figure*}[!htb]
    \centering
    \subfigure[Ellipsoid obstacle]{
    \label{fig:elip}
    \includegraphics[width=0.2\textwidth]{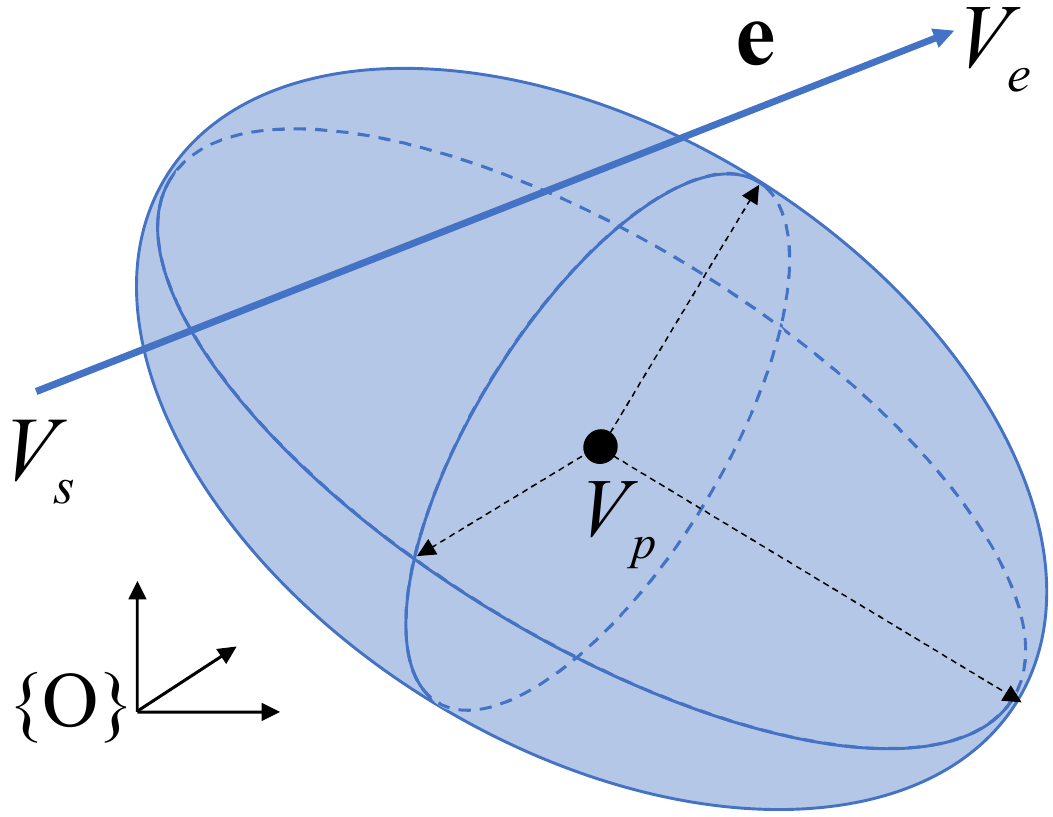}}~\hfil
	\subfigure[Unit sphere]{
    \label{fig:sphe}
    \includegraphics[width=0.17\textwidth]{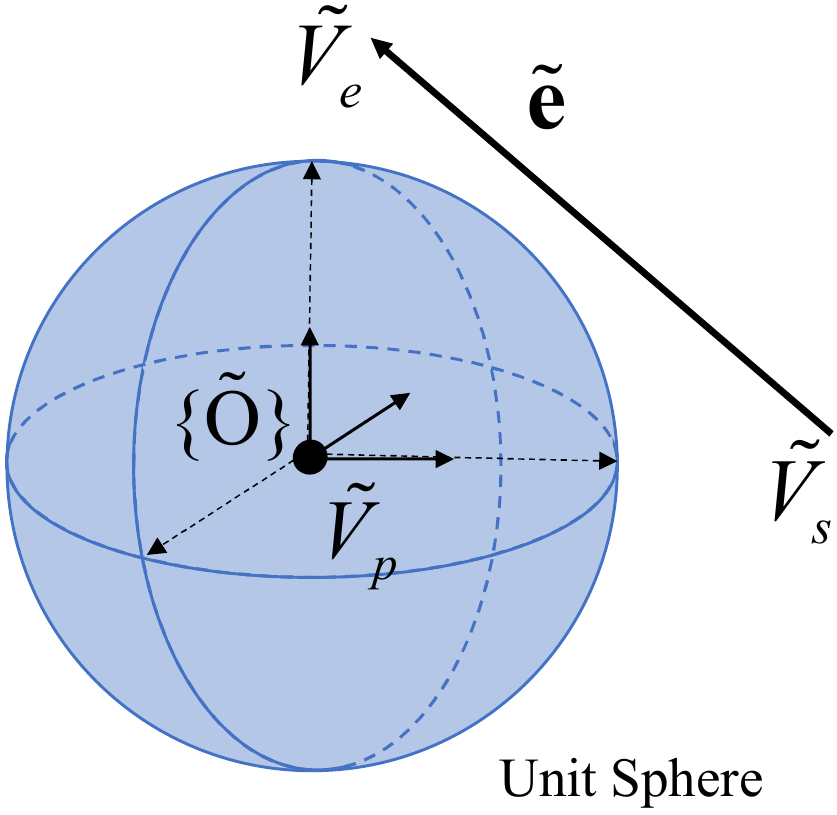}}~\hfil
    \subfigure[Cylinder obstacle]{
    \label{fig:clin}
    \includegraphics[width=0.19\textwidth]{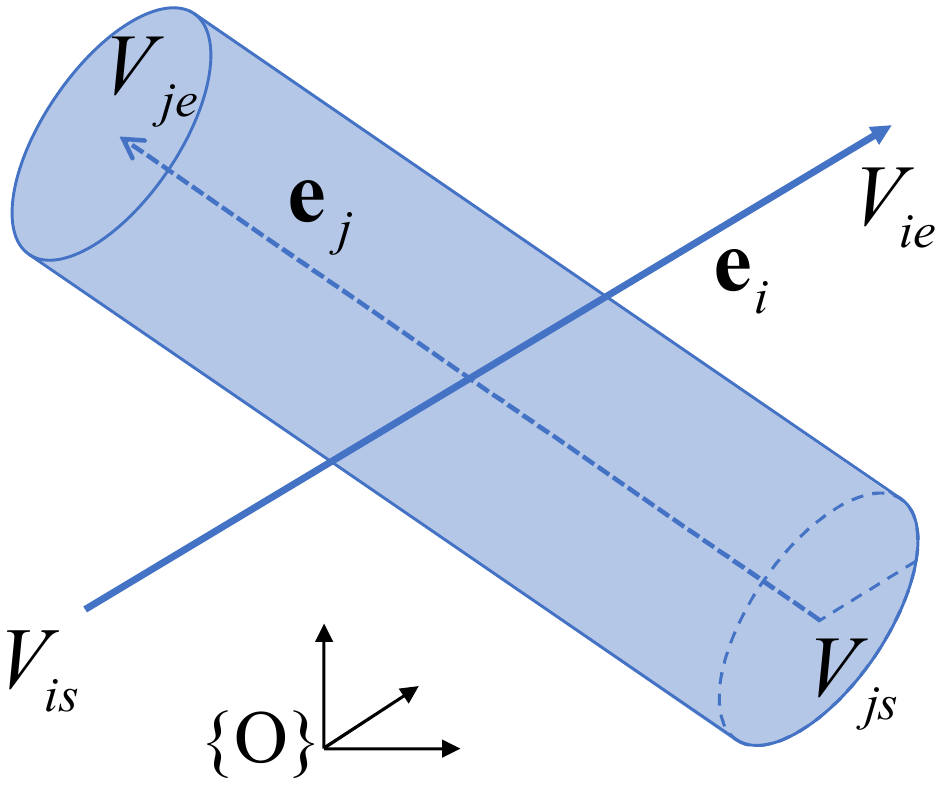}}~\hfil
    \subfigure[Polyhedron obstacle]{
    \label{fig:poly}
    \includegraphics[width=0.2\textwidth]{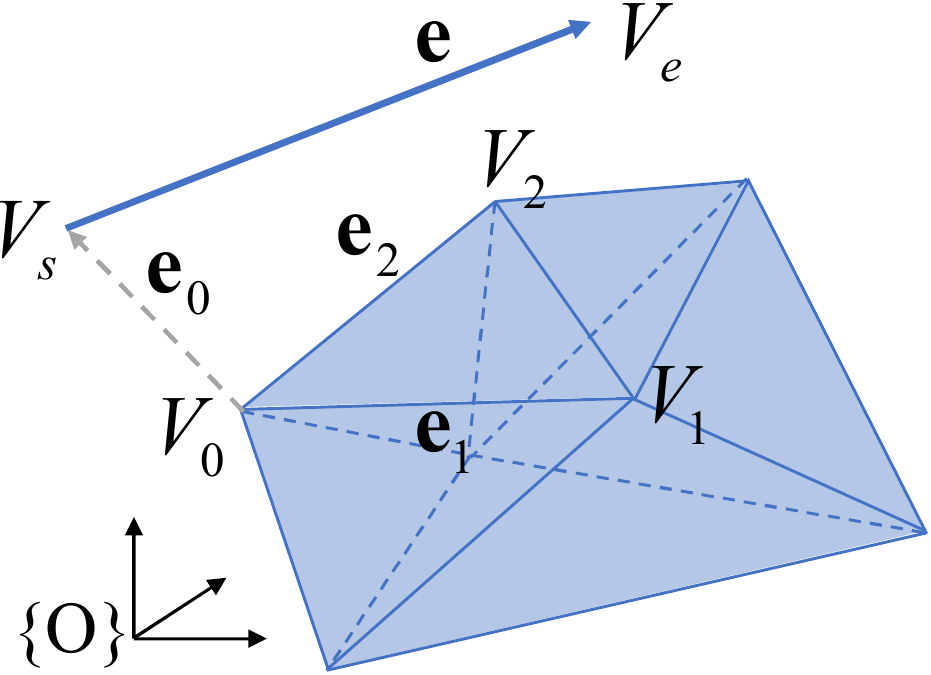}}~\hfil
    \vspace{-5pt}
    \caption{A family of obstacle models.}
    \vspace{-15pt}
    \label{fig:lotsOfObstacles}
\end{figure*}

\subsection{Collision Conditions for a Family of Obstacles}
Based on the discretized perception information from the real sensors, e.g., point clouds, occupancy grids, we would like to use a family of obstacle models to envelop these perceived obstacles in our CCD framework, such as \prettyref{fig:uav_elip}(a). Given the representative obstacle models, in this subsection, we will demonstrate that collision conditions can be transformed as a sequence of polynomial inequalities.

\subsubsection{Ellipsoid}
The ellipsoid in \prettyref{fig:elip} can be formulated as a group of points $\mathbf{x}$ in the world frame $\{O\}$ such that
\begin{align}\label{eq:def_ellipsoid}
    \mathcal{H} = \{\mathbf{x}: (\mathbf{x} - \mathbf{c})^T\mathbf{A}(\mathbf{x} - \mathbf{c}) = 1\},
\end{align}
where $\mathbf{c}:= \overrightarrow{OV_p}$ and $\mathbf{A}$ is a symmetric positive definite matrix, whose eigendecomposition is $\mathbf{A} = \mathbf{Q} \Lambda \mathbf{Q}^T$.
\SetKwFunction{AT}{AT}
It is known that the affine transformation (\AT), defined by
\begin{align}\label{eq:ellipAffine}
     \tilde{\mathbf{x}}  = \AT(\mathbf{x}):= \Lambda^{1/2} \mathbf{Q}^T(\mathbf{x} - \mathbf{c}),
\end{align}
can transform the ellipsoid $\mathcal{H}$ into the unit sphere $\mathcal{S} = \{\tilde{\mathbf{x}}: \tilde{\mathbf{x}}^T\tilde{\mathbf{x}}= 1\} $ centered at the origin in a new frame $\{\tilde{O}\}$, as displayed in \prettyref{fig:sphe}.
Since in Euclidean space, the affine transformation \prettyref{eq:ellipAffine} is a geometric transformation that preserves lines, then the $\tilde{\mathbf{e}}$, transformed from the edge $\mathbf{e}:= \overrightarrow{{V}_s{V}_e}$ in frame $\{O\}$ by \prettyref{eq:ellipAffine}, is still an edge in frame $\{\tilde{O}\}$.
As such, the collision case between edge and ellipsoid, as depicted in \prettyref{fig:elip}, can be transformed into the interference between edge and point with the safe distance $\underline{d} = 1$, as shown in \prettyref{fig:sphe}. Thus the collision condition becomes $\mathbf{dEP}(\tilde{\mathbf{e}},\tilde{O}) \leqslant 1$, as stated in \prettyref{sec:subsub_EP}.

Therefore, based on three conditions of \prettyref{eq:edge_pt}, firstly we can define the following items from $\mathbf{dEP}(\tilde{\mathbf{e}},\tilde{O})$:
\begin{align}\label{eq:u7ellip}
    &w_1 := 1- \|\tilde{\mathbf{p}}_s\|^2,~ w_2 := \tilde{\mathbf{p}}_s \cdot \tilde{\mathbf{e}}, \nonumber\\
    &w_3 := \|\tilde{\mathbf{e}}\|^2-\|\tilde{\mathbf{e}} \times \tilde{\mathbf{p}}_s\|^2,~ w_4 := -w_2,~ w_5 := w_2 + \tilde{\mathbf{e}} \cdot \tilde{\mathbf{e}},\nonumber\\
    &w_6 := 1- \|\tilde{\mathbf{p}}_e\|^2,~ w_7 := -w_5.
\end{align}
Here $\tilde{\mathbf{e}} := \overrightarrow{\tilde{V}_s\tilde{V}_e} = \AT(\mathbf{e})$, $\tilde{\mathbf{p}}_s := \overrightarrow{\tilde{O}\tilde{V}_s} = \AT(\mathbf{p}_s)$ and $\tilde{\mathbf{p}}_e := \overrightarrow{\tilde{O}\tilde{V}_e} = \AT(\mathbf{p}_e)$.
After that, according to \prettyref{eq:vertex_seg_t_3d}, it can be found that vectors of the vertices $\tilde{V}_s$, $\tilde{V}_e$ and the edge $\tilde{\mathbf{e}}$ in \prettyref{fig:sphe}, transformed from ${V}_s$, ${V}_e$ and $\mathbf{e}$ in \prettyref{fig:elip} by \prettyref{eq:ellipAffine}, can also be written as a set of polynomials with the same degrees, that is, 
\begin{align} \label{eq:vertex_seg_elli}
    \begin{split}
        &\mathbf{p}_s, \mathbf{p}_e, \tilde{\mathbf{p}}_s, \tilde{\mathbf{p}}_e \in 
        [P(t^{\overline{q}}), P(t^{\overline{q}}), P(t^{\underline{q}})]^T,\\
        &\mathbf{e}, \tilde{\mathbf{e}} \in 
        [P(t^{3p}), P(t^{3p}), P(t^{2p})]^T,\\
        &\overline{q} = \max(n, 3p), 
        \underline{q} = \max(n, 2p).
    \end{split}
\end{align}
After substituting $\tilde{\mathbf{p}}_s$, $\tilde{\mathbf{p}}_e$ and $\tilde{\mathbf{e}}$ in \prettyref{eq:vertex_seg_elli} into \prettyref{eq:u7ellip}, we could convert $w_1 \sim w_7$ into a set of univariate polynomials w.r.t. $t$. Therefore, based on  \prettyref{eq:edge_pt}, the collision conditions between the \textbf{E}dge $\tilde{\mathbf{e}}$ (suppose the $l$-th) and the \textbf{P}oint (origin) $\tilde{O}$, i.e., $\mathbf{dEP}(\tilde{\mathbf{e}},\tilde{O}) \leqslant 1$, can be rewritten as following polynomial inequalities w.r.t. $t$:
\begin{align}\label{eq:ineq_dEP}
    C_{EP}^l &= \{t\in[t_s,t_e]: C_1 \cup C_2 \cup C_3\}\\
    &C_1 = \{t: w_1 \geqslant 0, w_2 \geqslant 0\}, \notag\\
    &C_2 = \{t: w_3 \geqslant 0, w_4 > 0, w_5 > 0\}, \notag\\
    &C_3 = \{t: w_6 \geqslant 0, w_7 \geqslant 0 \}. \notag
\end{align}
Finally, due to $\varepsilon$ edges of the robot model, the collision conditions between the robot and the ellipsoid could be given as
$
    C_{Elip} = \bigcup_{l= 1}^{\varepsilon} C_{EP}^l .
$
It is worth noting that since the sphere is a special case of the ellipsoid \prettyref{eq:def_ellipsoid}, then $C_{Elip}$ works for CCD of the robot with the sphere obstacle as well.

\subsubsection{Cylinder}
The cylindrical object is defined by a constant vector $\mathbf{e}_j := \overrightarrow{V_{js}V_{je}}$ in the world frame $\{O\}$ with the given radius $\underline{d}$, as depicted in \prettyref{fig:clin}. Therefore, the interference of the edge $\mathbf{e}_i := \overrightarrow{V_{is}V_{ie}}$ to a cylinder can be detected using collision conditions of edge-edge in \prettyref{sec:sub_edgeedge}, i.e., $\mathbf{dEE}(\mathbf{e}_i, \mathbf{e}_j) \leqslant \underline{d}$. 

According to \prettyref{eq:vertex_seg_t_3d}, the vertex vectors $\mathbf{p}_{is} := \overrightarrow{OV_{is}}$, $\mathbf{p}_{ie} := \overrightarrow{OV_{ie}}$ and the  edge $\mathbf{e}_i$ of robot model can be rewritten as
\begin{align}
    \begin{split}
        &\mathbf{p}_{is}, \mathbf{p}_{ie} \in 
        [P(t^{\overline{q}}), P(t^{\overline{q}}), P(t^{\underline{q}})]^T,\\
        &\mathbf{e}_i \in 
        [P(t^{3p}), P(t^{3p}), P(t^{2p})]^T,\\
        &\overline{q} = \max(n, 3p),
        \underline{q} = \max(n, 2p).
    \end{split}
\end{align}
While the vertex vectors $\mathbf{p}_{js} := \overrightarrow{OV_{js}}$, $\mathbf{p}_{je} := \overrightarrow{OV_{je}}$ and the edge $\mathbf{e}_j$ are all constant vectors, determined by cylinder parameters.
Therefore, by definition of $\mathbf{e}_{ij} := \overrightarrow{V_{is}V_{js}} = \mathbf{p}_{js} - \mathbf{p}_{is}$ in \prettyref{eq:compact_edge_edge}, we can present $\mathbf{e}_{ij}$ as
\begin{align}\label{eq:eij_3d}
    \mathbf{e}_{ij} \in [ P(t^{\overline{q}}), P(t^{\overline{q}}), P(t^{\underline{q}})]^T.
\end{align}
Substituting above $\mathbf{e}_i$, $\mathbf{e}_j$ and \prettyref{eq:eij_3d} into \prettyref{eq:4itemsEE}, we obtain
\begin{align}\label{eq:ntintjd_t}
    \begin{split}
        &u_0 = \det\mathbf{M} \in P(t^{6p}),\\
        &u_1 \in P(t^{3p+\overline{q}}),~
         u_2 \in P(t^{6p+\overline{q}}),~
         u_3 \in P(t^{3p+\overline{q}}).
    \end{split}
\end{align}
If we substitute \prettyref{eq:ntintjd_t} into \prettyref{eq:dEE}, then the first collision condition between two edges can be reformulated as following polynomial inequalities w.r.t. time $t$:
\begin{align}\label{eq:ineq_dEE_1}
    C_1 = \{t\in[t_s,t_e]:& u_1 \geqslant 0,\ u_2 \geqslant 0,\ u_0 - u_1 \geqslant 0, \notag \\
    &u_0 - u_2 \geqslant 0,\ \underline{d}^2 u_0 - u_3^2 \geqslant 0\}.
\end{align}
It is worth noting that \prettyref{eq:ineq_dEE_1} holds because there is an underlying equation that is $u_0 = \|\mathbf{e}_i \times \mathbf{e}_j\|^2 > 0$.
From the second to fifth conditions in \prettyref{eq:dEE}, they could also be rewritten as a bunch of polynomial inequalities w.r.t. $t$. Taking the second one as an example, it becomes
\begin{align}
     C_2 = \{t\in[t_s,t_e]:& -u_1 >0,\ u_2 \geqslant 0, u_0 - u_2 \geqslant 0,\notag\\
     &\mathbf{dEP}(\mathbf{e}_j,V_{is}) \leqslant \underline{d}\},\nonumber
\end{align}
where $\mathbf{dEP}(\mathbf{e}_j,V_{is}) \leqslant \underline{d}$ can be reformulated as polynomial inequalities by \prettyref{eq:ineq_dEP} as well. For the last four conditions in \prettyref{eq:dEE}, it is straightforward to transfer them in polynomial inequality forms, e.g., the sixth condition is transformed as
\begin{align}
        C_6 = \{t\in[t_s,t_e]: &-u_1 >0, -u_2>0, \underline{d}^2-\|\mathbf{e}_{ij} \|^2 \geqslant 0\}.\nonumber
\end{align}
So, all collision conditions between the \textbf{E}dge $\mathbf{e}_i$ and the cylinder (\textbf{E}dge) $\mathbf{e}_j$, i.e., $\mathbf{dEE}(\mathbf{e}_i, \mathbf{e}_j) \leqslant \underline{d}$, can be summarized as
$C_{EE}^i = \{t\in[t_s,t_e]: \cup_{k= 1}^{9} C_k\}$.
Finally, the collisions between the robot with $\varepsilon$ edges and the cylinder are presented as
$
    C_{Clin} = \bigcup_{i= 1}^{\varepsilon} C_{EE}^i .
$

\subsubsection{Polyhedron}
For more general obstacles, it is convectional to approximate them by triangle meshes \cite{botsch2010polygon}, where it contains a group of triangular surfaces connected by their common edges or vertices, as shown in \prettyref{fig:poly}. 

Similarly, based on \prettyref{eq:vertex_seg_t_3d}, we can transform the vectors of vertices $V_s$, $V_e$ and the edge $\mathbf{e}$ as
\begin{align} \label{eq:vertex_seg_trimesh}
    \begin{split}
        &\mathbf{p}_s, \mathbf{p}_e \in 
        [P(t^{\overline{q}}), P(t^{\overline{q}}), P(t^{\underline{q}})]^T, \\
        &\mathbf{e} \in 
        [P(t^{3p}), P(t^{3p}), P(t^{2p})]^T,\\
        &\overline{q} = \max(n, 3p),
        \underline{q} = \max(n, 2p).
    \end{split}
\end{align}
Due to the constant vector $\overrightarrow{OV_{0}}$ and by definition of $\mathbf{e}_0 := \overrightarrow{OV_s}-\overrightarrow{OV_{0}} = \mathbf{p}_s - \overrightarrow{OV_{0}}$ in \prettyref{eq:compact_edge_tri}, we can rewrite $\mathbf{e}_0$ as three polynomial equations w.r.t. $t$:
\begin{align}\label{eq:e0_3d}
    \mathbf{e}_0 \in [P(t^{\overline{q}}), P(t^{\overline{q}}), P(t^{\underline{q}})]^T.
\end{align}
Substituting $\mathbf{e}$ in \prettyref{eq:vertex_seg_trimesh} and $\mathbf{e}_0$ in \prettyref{eq:e0_3d}, as well as constant vectors $\mathbf{e}_1$, $\mathbf{e}_2$, into  \prettyref{eq:4itemsTri}, it yields
\begin{align}\label{eq:nknk1nk2d_t}
    \begin{split}
        &v_3 = \det\mathbf{Q} \in P(t^{3p}),\\
        &v_0 \in P(t^{\overline{q}}),~~
        v_1 \in P(t^{3p+\overline{q}}),~~
        v_2 \in P(t^{3p+\overline{q}}).
    \end{split}
\end{align}
Furthermore, substituting \prettyref{eq:nknk1nk2d_t} into \prettyref{eq:cond_segTri}, the conflict conditions for the $\mathbf{E}$dge (suppose the $l$-th) and the $\mathbf{T}$riangle (suppose the $m$-th) can be reformulated as following polynomial inequalities w.r.t. $t$:
\begin{align}
    & C^{lm}_{ET} = \{t\in [t_s, t_e]: C_{+} \cup C_{-} \} \notag \\
    C_{+} = \{t:\ &v_3 >0, v_0 \geqslant 0, v_3 - v_0 \geqslant 0, v_1 \geqslant 0, v_2 \geqslant 0, \notag\\
    &v_3 - (v_1 + v_2) \geqslant 0 \} \notag\\
    C_{-} = \{t:\ &v_3 <0, v_0 \leqslant 0, v_3 - v_0 \leqslant 0, v_1 \leqslant 0, v_2 \leqslant 0, \notag\\
    &v_3 - (v_1 + v_2) \leqslant 0 \} \notag
\end{align}
At last, because of $\varepsilon$ edges of the model and total $\kappa$ triangles of the obstacle, the collision conditions for robot and polyhedron can be expressed as:
$
    C_{Poly} = \bigcup_{l= 1}^{\varepsilon}\bigcup_{m= 1}^{\kappa} C^{lm}_{ET}.
$

\section{Implementation Details}
\label{sec:application}
\subsection{Coefficient Determination Method of Polynomials}
In this subsection, we will demonstrate a general numerical approach to determine the coefficients of polynomials no matter which collision conditions they are. With the knowledge of polynomial $u$ with its degree $r$, it can be explicitly written as
\begin{align}
    u(t) = a_{r}t^{r} + a_{r-1}t^{r-1} + \cdots + a_{1}t + a_{0},
\end{align}
where $r+1$ unknown coefficients could be solved by taking $r+1$ unique samples of $t \in [t_s, t_e]$ (denoted by $t_i,~i= 1,\cdots,r+1$). Thus, its coefficients are solved using
\begin{align}
    \begin{bmatrix}
        a_{r} \\ a_{r-1} \\ \vdots \\ a_{0}
    \end{bmatrix}=
    \begin{bmatrix}
        t_1^r &t_1^{r-1} &\cdots &t_1 &1\\
        t_2^r &t_2^{r-1} &\cdots &t_2 &1\\
        \vdots & & & & \vdots\\
        t_{r+1}^r &t_{r+1}^{r-1} &\cdots &t_{r+1} &1\\
    \end{bmatrix}^{-1}
    \begin{bmatrix}
        u(t_1) \\ u(t_2) \\ \vdots \\ u(t_{r+1})
    \end{bmatrix}.\nonumber
\end{align}

This approach is able to readily determine the polynomial formulations in above sections considering each polynomial degree has been given clearly. In this way, we can omit tedious polynomial estimation, i.e.,  \prettyref{eq:sincosFit3d}, and ignore the complicated symbolic substitution process, such as substituting polynomial vectors $\mathbf{e}_i,\mathbf{e}_{ij}$ into \prettyref{eq:4itemsEE} to obtain \prettyref{eq:ntintjd_t}.

\subsection{Efficient Method to Solve Polynomial Inequalities}
Even if there exist some computationally efficient solvers to find roots of the polynomial equation, such as \cite{yang2020applied}, we will introduce a simple yet efficient roots solving algorithm to reduce the computation cost further.

The key idea lies in the utilization of the Sturm's Theorem \cite{Basu2006} before directly solving roots of polynomials, and the whole process is presented in \prettyref{algo:sturm}. Specifically, given a set of polynomial inequalities, e.g., \prettyref{eq:ineq_dEE_1}, and denoted by $\mathbf{G}(t) = \{t\in [t_s,t_e]: g_1(t) \geqslant 0,\ g_2(t) \geqslant 0, \cdots,\ g_k(t) \geqslant 0\}$, we will check whether each polynomial $g_i(t), i=1,\cdots,k$ is satisfied over all $t \in [t_s,t_e]$.

\begin{algorithm}[htbp]
    \SetKwFunction{add}{add($i$)}
    \SetKwFunction{Signa}{Sign($g_i(t_s)$)}
    \SetKwFunction{Signb}{Sign($g_i(t_e)$)}
	\caption{Efficient method to solve the polynomial inequalities} 
	\label{algo:sturm}
	\SetKwInOut{Input}{Input}\SetKwInOut{Output}{Output}
    \SetKwData{lower}{lower}\SetKwData{upper}{upper}\SetKwData{result}{result}
	    \Input{$g_1(t)\geqslant 0, \cdots, g_k(t)\geqslant 0, \forall t\in[t_s,t_e]$} 
	    \Output{resulting intervals $\mathbf{G}(t)$}
	    $L \leftarrow \varnothing$\;
		\For{$i\leftarrow 1$ \KwTo $k$}
		    {
		    $\#Roots \leftarrow g_i(t)=0$ \emph{over} $(t_s, t_e]$ \emph{by Sturm's Theorem}\;\label{alli:numRts}
	        \lIf{$\#Roots > 0$}{$L.$\add}
            \Else{
                \lIf{\Signa$>0$}{\bf continue \label{alli:continueA}}
                \ElseIf{\Signa$<0$}{$\bf return\ \mathbf{G}(t) \leftarrow \varnothing$ \label{alli:returnEmptyA}\;}
                \Else{
                    \lIf{\Signb$>0$}{\bf continue \label{alli:continueB}}
                    \lElse{$\bf return\ \mathbf{G}(t) \leftarrow \varnothing$ \label{alli:returnEmptyB}}
                    }
                 }
		    }
	    \lIf{$L$ is empty}
	        {$\mathbf{G}(t) \leftarrow [t_s, t_e]$}
        \lElse{$\mathbf{G}(t) \leftarrow$ \emph{solve} $g_j(t) \geqslant 0, \forall j \in L, \forall t \in [t_s, t_e]$\label{alli:roots}}
        {\bf return\ $\mathbf{G}(t)$}
\end{algorithm}
\vspace{-18pt}

\section{Experiments}
\label{sec:experiment}
\subsection{Collision Checking of Quadrotor}
Due to the differential flatness for quadrotor \cite{zhou2014vector}, we can uniquely express the trajectory in space of flat outputs, where a possible choice of flat outputs is $[x,y,z,\psi]^T$. Then based on its dynamic property, the $\phi$ and $\theta$ could be derived by flat outputs and their time derivative as well: 
$
\phi = \text{sin}^{-1}(\frac{\sin\psi \ddot{x} - \cos\psi \ddot{y}}{\sqrt{\ddot{x}^2+\ddot{y}^2+(g+\ddot{z})^2}})$, $
\theta = \text{tan}^{-1}(\frac{\cos\psi \ddot{x} + \sin\psi \ddot{y}}{\ddot{z} + g}),$
where $g$ is the gravitational acceleration.
So, given one piece polynomial trajectories of $x(t), y(t), z(t)$ and $\psi = 0$ at all times, we could well present each entry of rotation matrix \prettyref{eq:rotMat3d} by \prettyref{eq:sincosFit3d} in the polynomial forms. 
In this case, we set the same end condition for quadrotor $\mathbf{s}_{g} = [600, 650, 700,0,0,0,0,0,0]$ accounting for $x,y,z,\dot{x},\dot{y},\dot{z},\ddot{x},\ddot{y},\ddot{z}$, respectively.

\subsubsection{An Ellipsoid}
According to \prettyref{eq:def_ellipsoid}, an ellipsoid obstacle $\mathcal{H}$ is defined by $\mathbf{c}$ and $\mathbf{A}$ in \prettyref{fig:uav_elip}(b).
Given the start condition $\mathbf{s}_{s} =[3, 3, 3, 3.2, 0, 2, 0, 0, 0]\times100$ and the end condition $\mathbf{s}_g$, we generate a minimum snap trajectory \cite{mellinger2011minimum} T1 with $7$-th degree polynomials w.r.t. $t \in [0, 3]$ as
\begin{align}\label{eq:uavTraj1}
    \begin{split}
        x &= -0.02t^7+0.75t^6\cdots+320t+300,\\
        y &= -0.07t^7+1.32t^6\cdots+111.11t^3+300,\\
        z &= +0.01t^7-0.51t^6\cdots+200t+300.
    \end{split}
\end{align}
So, $n= 7$ from \prettyref{eq:xyzPsiPoly3D}.
Taking the first edge in \prettyref{fig:uav3d} as an example, the $w_1 \sim w_3$ and $w_6$ in \prettyref{eq:u7ellip} can be derived as
\begin{align}
    w_1 = &-1.27\times10^{-9}t^{16}+1.83\times10^{-8}t^{15}\cdots
           -5.67, \nonumber\\
    w_2 = &\ 3.71\times10^{-8}t^{12}-1.37\times10^{-6}t^{11}\cdots
           + 0.78, \nonumber\\
    w_3 = &-1.35\times10^{-14}t^{24}+6.61\times10^{-13}t^{23} \cdots
           -1.21, \nonumber\\
    w_6 = & -1.27\times10^{-9}t^{16}+1.83\times10^{-8}t^{15}  \cdots
           - 7.54. \nonumber
\end{align}
It is worth noting that since $\psi$ is constant, $\mathbf{e}, \tilde{\mathbf{e}} \in [P(t^{2p}), P(t^{2p}), P(t^{2p})]^T$, $\overline{q} = \underline{q} = \max(n, 2p) = 8$ with $p=4$ in \prettyref{eq:vertex_seg_elli}. So based on \prettyref{eq:ineq_dEP}, the collision interval between the first edge (i.e., $l = 1$) and the ellipsoid is computed as $C_{EP}^1 = \{t: [0.6959, 1.4420]\}$. After checking total $\varepsilon =12$ edges, the collision intervals for the quadrotor and the ellipsoid is calculated by 
$
    C_{Elip} = \bigcup_{l= 1}^{12} C_{EP}^l = \{t: [0.6959, 1.6347]\},\nonumber
$
which is shown in red curve in \prettyref{fig:uav_elip}(b). Here the time of impact, i.e., $t = 0.6959s$, has been exactly computed.  In addition, we demonstrate the poses of quadrotor at two end points of the $C_{Elip}$, which refer to the starting and terminal situations of collision, respectively.

\subsubsection{Cylinders and Polyhedra}
Several cylindrical and polyhedral obstacles are randomly located in the environment and seven minimum snap trajectories T1 $\sim$ T7 over $3s$ are generated according to various initial velocities at start conditions $\mathbf{s}_s$, as presented in \prettyref{fig:uav_obs}. The resulting collision intervals w.r.t. time $t$ are listed in  \prettyref{tab:collIntv2cases}.

In \prettyref{fig:uav_clin}, we can find that the quadrotor along T4 is seriously affected by cylinders, resulting in three collision intervals, whereas it can safely fly through T7. In \prettyref{fig:uav_triM_L} and \prettyref{fig:uav_triM_S}, we test
quadrotors with different sizes given in \prettyref{tab:collIntv2cases} in the same polyhedral environment. It is shown that with the size decreasing, the small quadrotor is able to pass through obstacles by the trajectory T5, along which the large quadrotor has collisions.
It is worth mentioning that there are two collision intervals for the small quadrotor with T3, as shown in the $3$-rd column of \prettyref{tab:collIntv2cases} and \prettyref{fig:uav_triM_S}. This is because in our framework, we directly leverage the intrusion conditions between edge and triangle without considering the safe distance,  thus it is possible to provide collision free intervals within the obstacle interior. But it is acceptable in real applications due to the ToI determined by our approach. Furthermore, based on the knowledge of the environment, we also could ignore the resulting collision free intervals within the obstacle interiors.

\subsubsection{Comparison}
\begin{figure}[!htb]
    \centering
    \subfigure[Cylinders]{
    \label{fig:uav_clin}
    \includegraphics[width=0.15\textwidth]{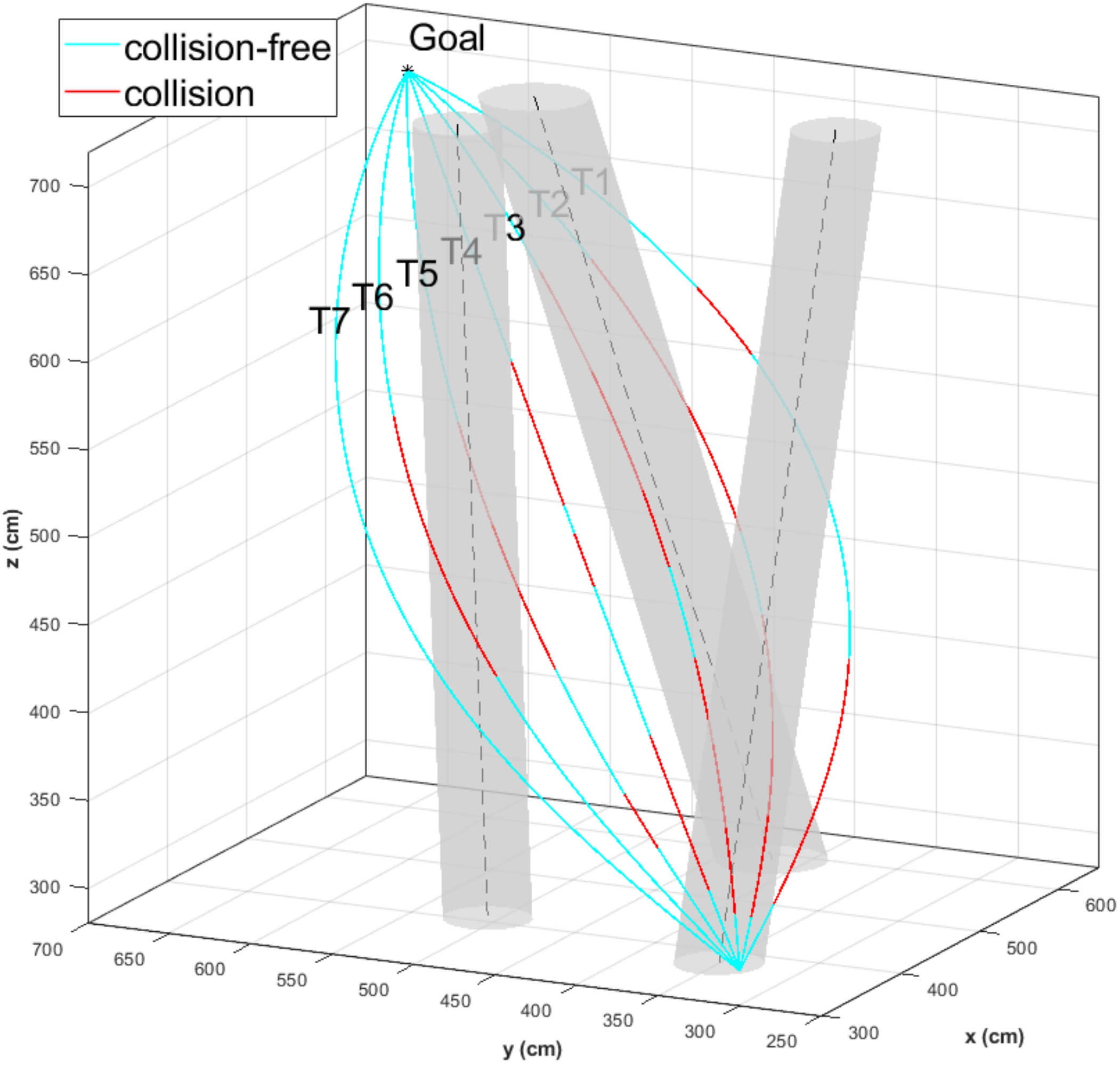}}~\hfil
    \subfigure[Polyhedra]{
    \label{fig:uav_triM_L}
    \includegraphics[width=0.15\textwidth]{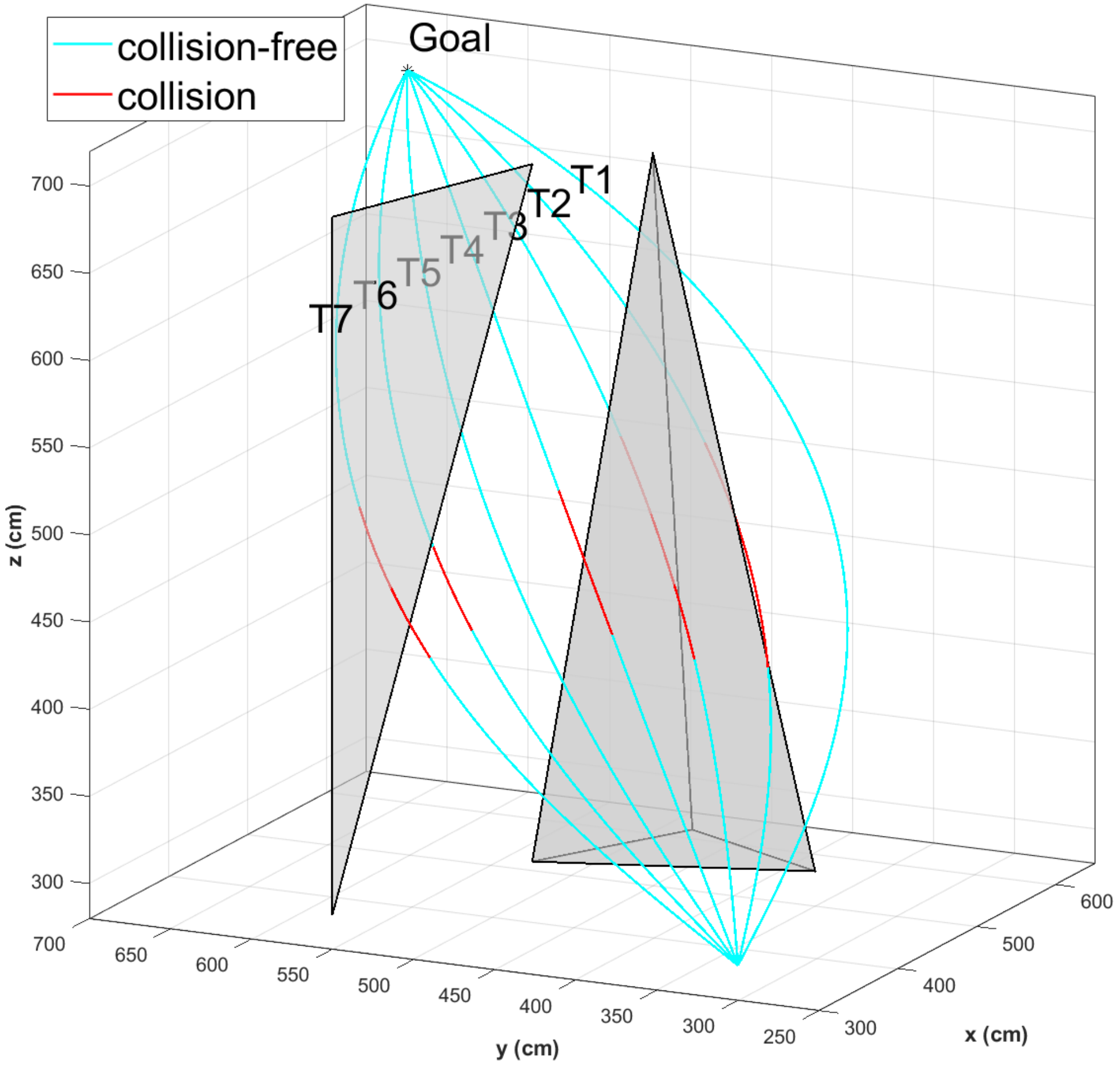}}~\hfil
    \subfigure[Polyhedra]{
    \label{fig:uav_triM_S}
    \includegraphics[width=0.15\textwidth]{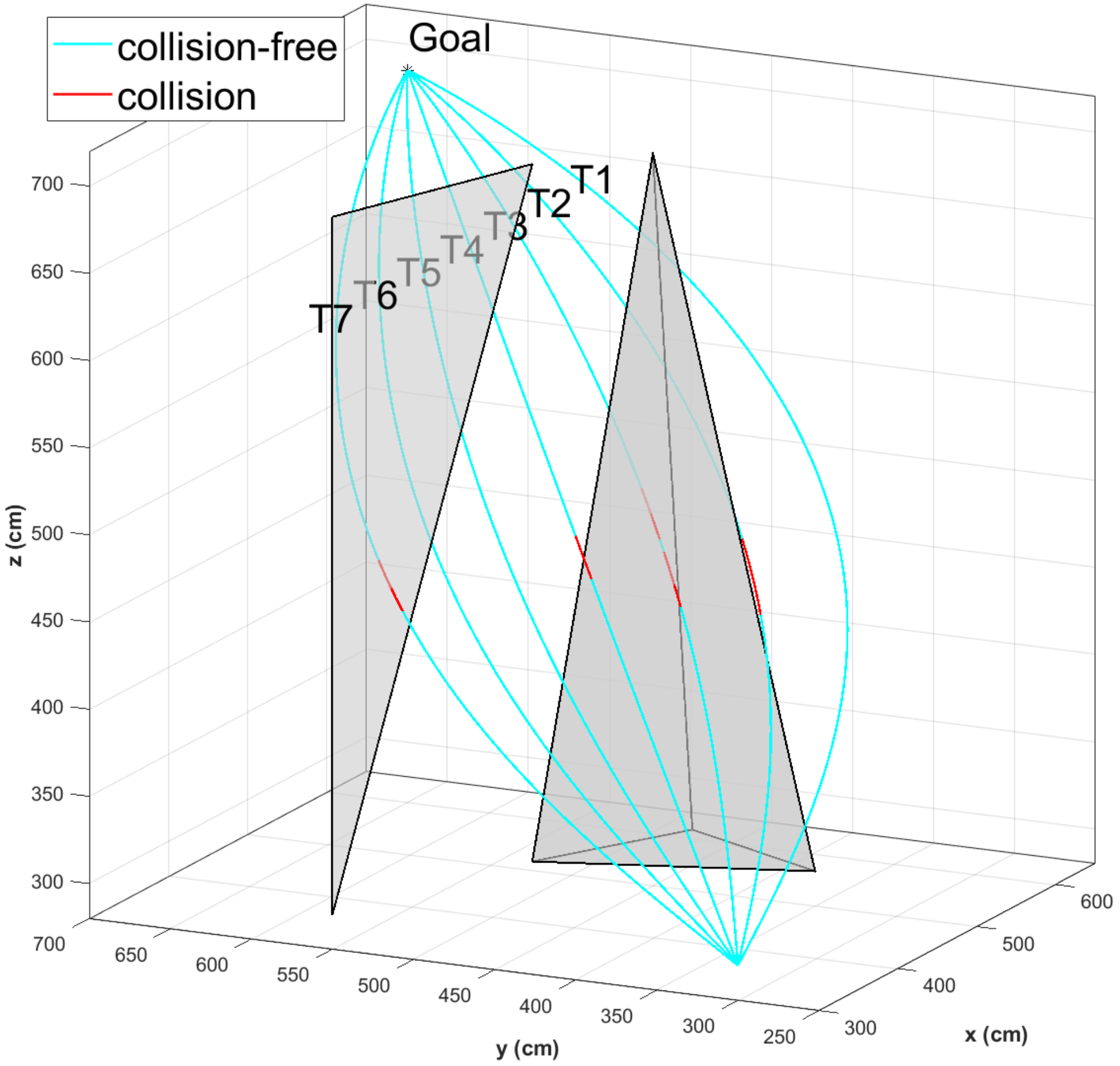}}
    \vspace{-10pt}
    \caption{CCD between a quadrotor with various obstacles, which are transparent to get better display. In (a) and (b), the size of quadrotor is $60 \times 60 \times 20\ cm$, whereas a smaller quadrotor, $20 \times 20 \times 20\ cm$, is tested in (c).}
    \vspace{-10pt}
    \label{fig:uav_obs}
\end{figure}

\prettyref{tab:timeCost} demonstrates a direct comparison of time efficiency between the proposed method and the well-accepted point-wise approach with GJK\cite{ref_ericson2004real} in \prettyref{fig:uav_triM_L}. All results are generated using a computer with Intel\textsuperscript{\tiny\textregistered} Core\texttrademark~i7-6500U CPU.

We generate $91$ minimum snap trajectories with duration of $3s$, $5s$ and $10s$, respectively. For the case of  trajectory over $3s$, the resolution of results from our method is more than $0.001s$, as listed in \prettyref{tab:collIntv2cases}, which is expected due to the accuracy of the numerical polynomial roots solver. Nevertheless, its time cost ($0.202s$) is less than that of the point-wise approach with $\Delta t = 0.01s$ ($0.268s$). If time step gets finer, such as $\Delta t = 0.001s$, the point-wise approach takes $2.735s$ to check the collision for each trajectory, which is over 10 times slower than our method. Furthermore, it is obvious that the time efficiency of our method further outperforms the point-wise technique for longer trajectories.

\begin{table}[htbp]
	\caption{Collision Intervals (Unit: $s$)}
	\vspace{-5pt}
	\centering 
	\resizebox{0.48\textwidth}{!}{
	\begin{tabular}{c | c | c | c} 
		\hline\hline 
		\multirow{2}{*}{Traj.} &  $C_{Clin}$ &   $C_{Poly}$ &    $C_{Poly}$ \\
		& (size: $60\times60\times20\ cm$) & ($60\times60\times20$) & ($20\times20\times20$)\\
		\hline 
		T1 & $[0.1375, 0.6433] \cup [1.3259, 1.5071]$ & $\varnothing$ & $\varnothing$\\
		\hline
		T2 & $[0.1083, 0.7165] \cup [0.8911, 1.5656]$ & $[0.6021, 1.0833]$ & $[0.7095, 0.8705]$\\
		\hline
		\multirow{2}{*}{T3} & \multirow{2}{*}{$[0.1127, 0.6265] \cup [0.8131, 1.5279]$} & \multirow{2}{*}{$[0.6159, 1.0944]$} & $[0.7238, 0.8400]$ \\
		& & & $\cup [0.8657, 0.9774]$\\
		\hline
		\multirow{2}{*}{T4} & $[0.1620, 0.4751]\cup [0.7808, 0.8933]$   & \multirow{2}{*}{$[0.6742, 0.9823]$} &  \multirow{2}{*}{$[0.7906, 0.8828]$} \\
		& $\cup [0.9552, 1.2878]$ & &\\
		\hline
		T5 & $[0.2563, 0.3692] \cup [0.6283,1.1637]$   & $\varnothing$ & $\varnothing$\\
		\hline
		T6 & $[0.6391, 1.2124]$   &  $[0.7307, 0.9113]$ &  $\varnothing$\\       
		\hline
		T7 & $\varnothing$  & $[0.7120, 1.0472 ]$ & $[0.8151, 0.9269]$\\
		\hline
	\end{tabular}
	}
	\label{tab:collIntv2cases}
	\vspace{-15pt}
\end{table}

\begin{table}[htbp]
    \caption{Average Computation Cost for Collision Checking with Polyhedral Obstacles (Unit: $s$)}
	\centering
    \begin{tabular}{c | c| c| c| c| c| c}
    \hline\hline 
	{Time Step} & \multicolumn{2}{c|}{Traj. $t \in [0, 3]$} & 
	\multicolumn{2}{c|}{Traj. $t \in [0, 5]$} & \multicolumn{2}{c}{Traj. $t \in [0, 10]$} \\
	\cline{2-3} \cline{4-5} \cline{6-7}
	$\Delta t$ & Our & GJK & Our  & GJK & Our & GJK \\
	\hline
	0.1 & \multirow{3}{*}{0.202} & 0.029 & \multirow{3}{*}{0.195} & 0.048 & \multirow{3}{*}{0.167} & 0.097 \\
	\cline{1-1}\cline{3-3}\cline{5-5}\cline{7-7}
	0.01 & & 0.268 & & 0.466 & & 0.922\\
	\cline{1-1}\cline{3-3}\cline{5-5}\cline{7-7}
	0.001 & & 2.735 & & 4.631 & & 9.540 \\
	\hline
    \end{tabular}
    \vspace{-10pt}
    \label{tab:timeCost}
\end{table}

\subsection{Collision Checking of CDPR}
\begin{figure}[htbp]
    \centering
    \subfigure[CDPR model]{
    \label{fig:cdpr_model}
    \includegraphics[width=0.2\textwidth]{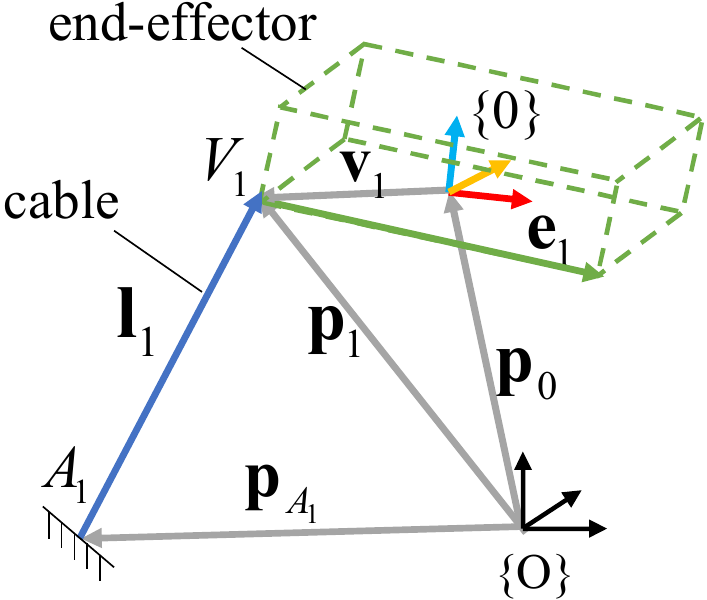}}~\hfil
    \subfigure[Two collision intervals]{
    \label{fig:oneColi}
    \includegraphics[width=0.2\textwidth]{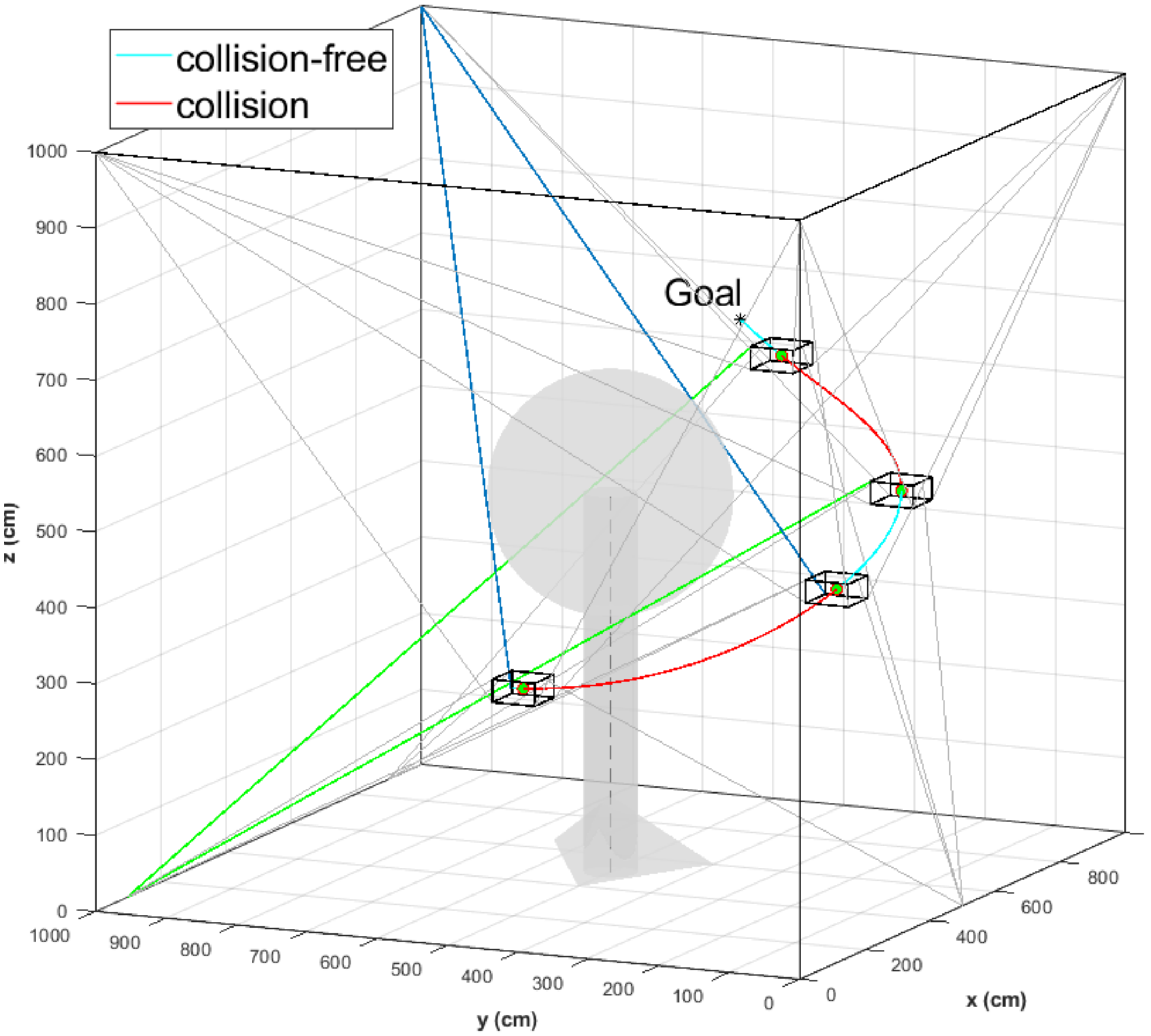}}~\vfil
    \subfigure[Without collision]{
    \label{fig:oneNoColi}
    \includegraphics[width=0.2\textwidth]{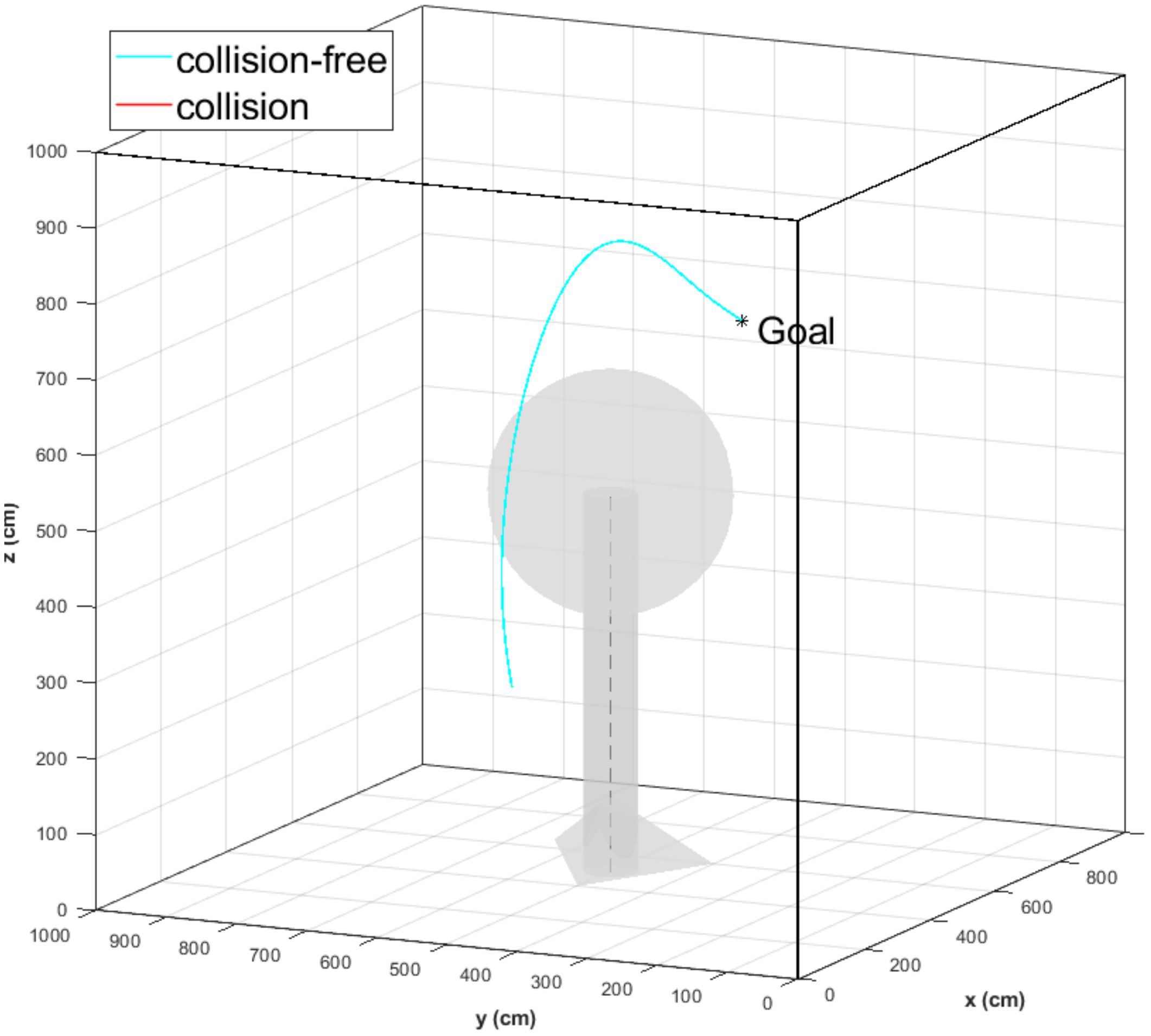}}~\hfil
    \subfigure[Collision free workspace]{
    \label{fig:ifw}
    \includegraphics[width=0.2\textwidth]{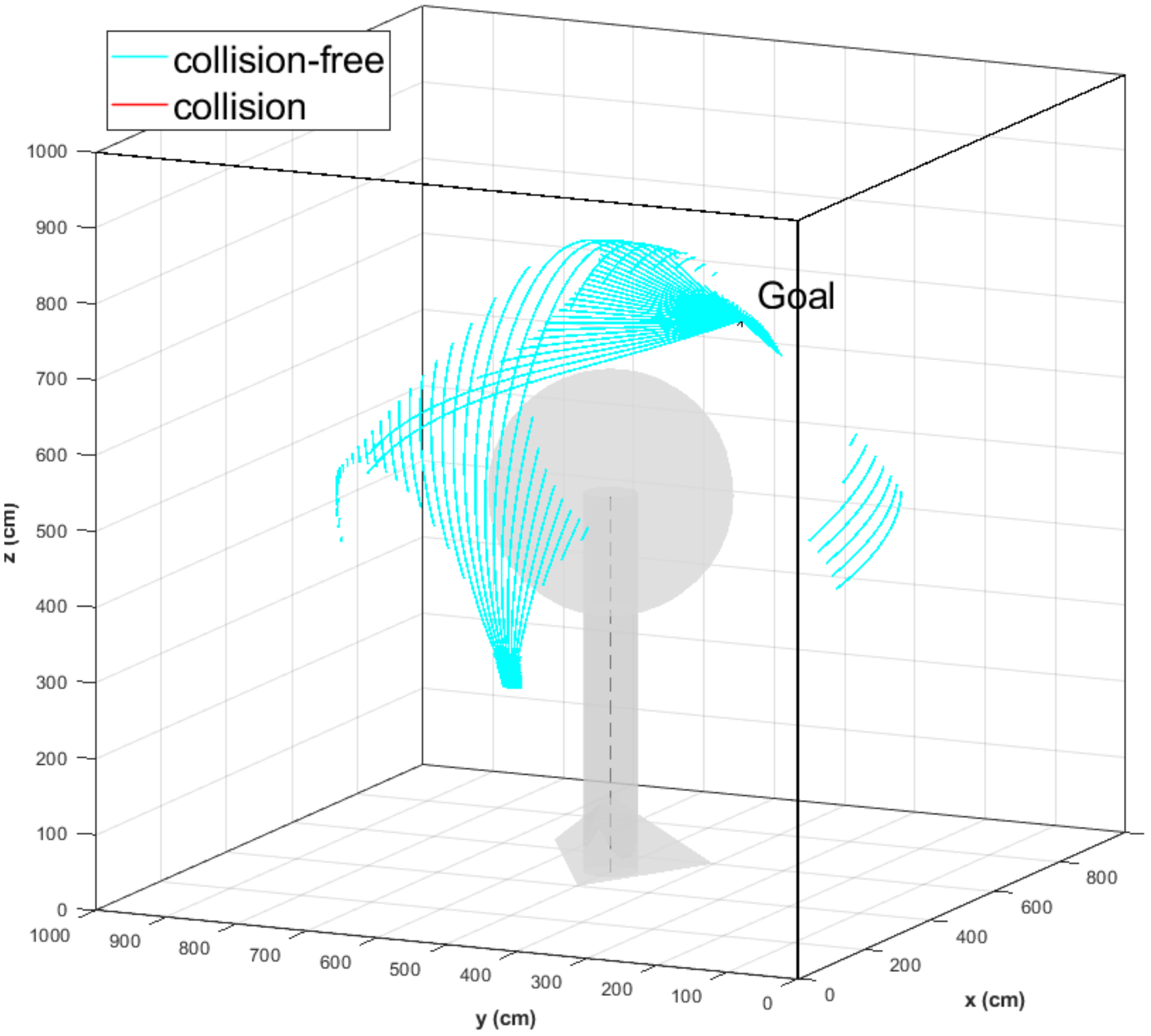}}
    \vspace{-10pt}
    \caption{CDPR model and CCD for cables with the tree. The collision intervals in (b) are $t \in [0.08, 3.16] \cup [5.20, 8.82]$.}
    \vspace{-10pt}
    \label{fig:cdpr_tree}
\end{figure}

The CDPR is another type of mobile platforms, which is driven by multiple cables attached on the same end-effector, as shown in \prettyref{fig:cdpr_model}. The collision of cables with environments seriously affects the workspace of the CDPR, thus collision detection for cables should be well studied \cite{zhang2019efficient}. 
In this experiment, there is a tree within the workspace of the CDPR, so we first model the tree as a combination of a sphere, a cylinder and a polyhedron, which provides more compact representation for the obstacle.

Unlike the quadrotor that could be enclosed as a bounding box, the cables can only be modeled as edges. In \prettyref{fig:cdpr_model}, the vector of cable $\mathbf{l}_1$ is determined by
$
    \mathbf{p}_1 = \mathbf{p}_{0} + R~\mathbf{v}_1,~
    \mathbf{l}_1 = \mathbf{p}_1 - \mathbf{p}_{A_1},
$
where $\mathbf{p}_1$ is same as \prettyref{eq:vertex2D} and $\mathbf{p}_{A_1}$ is a constant vector accounting for an anchor point of cable in the world frame $\{O\}$. Since the orientation motion of CDPR can be defined separately from translation trajectory, thus, in this subsection, we just consider the constant orientation movement, i.e., $\phi = \theta = \psi = 0$. In other words, the rotation matrix $R$ is a constant matrix, and $\mathbf{p}_{0}$ is defined by 3 polynomial equations, i.e., $x(t), y(t), z(t) \in P(t^n)$.
So, it is straightforward to find that the vectors of vertex and cable can be rewritten as polynomial equations:
$
    \mathbf{p}_1, \mathbf{l}_1 \in [P(t^n), P(t^n), P(t^n)]^T.
$
As such, given polynomial trajectories, the collision between cables and environment can be checked by our CCD framework. 

In \prettyref{fig:oneColi}-\subref{fig:ifw}, we test lots of Bezier paths, as one of polynomial curves, and check the CCD between cables and the tree.
It can be found that there two disjoint collision intervals on the path in \prettyref{fig:oneColi}. Here we depict the robot poses at endpoints of collision intervals, and further highlight cables that cause these collisions. 
In addition, we also find a safe path that passes through the ``crown'' overhead, as described in \prettyref{fig:oneNoColi}. 
To exploit more working areas of the CDPR, we demonstrate its interference free workspace in the form of trajectories in \prettyref{fig:ifw}, where we test $51$ paths and only show the collision free intervals. From the result, we can see there exist two separate areas, where the smaller one on the right does not connect the start point and goal, whereas the larger part on the left can be selected safely for CDPR motion.

\subsection{The Navigation System for AGV}
Given the polynomial trajectory $x(t), y(t) \in P(t^n), t\in[t_s, t_e]$ for AGV, then its orientation $\alpha$ is determined by the tangent direction at point $(x(t), y(t))$ due to the nonholonomic constraint, i.e.,
$
    \alpha = \text{tan}^{-1}(\frac{dy/dt}{dx/dt}).
$
As such, we can well estimate $\sin\alpha$ and $\cos\alpha$ in the rotation matrix by polynomial approximation method, similar to \prettyref{eq:sincosFit3d}. 

\begin{figure}
    \centering
    \subfigure[Sparse polygonal obstacles]{
    \label{fig:sparseObs2}
    \includegraphics[width=0.2\textwidth]{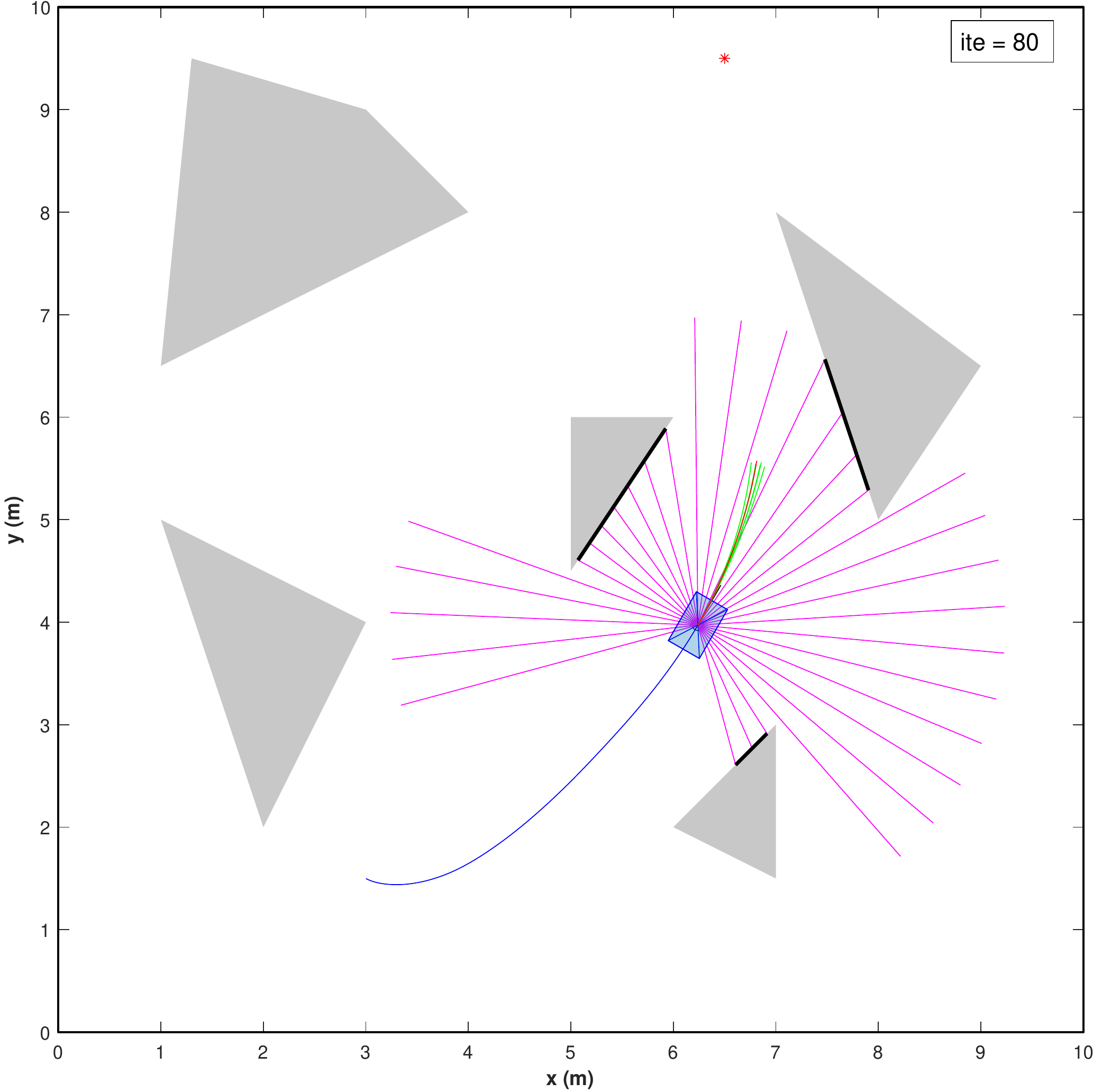}}~\hfil
	\subfigure[A narrow corridor]{
    \label{fig:narrowTube}
    \includegraphics[width=0.2\textwidth]{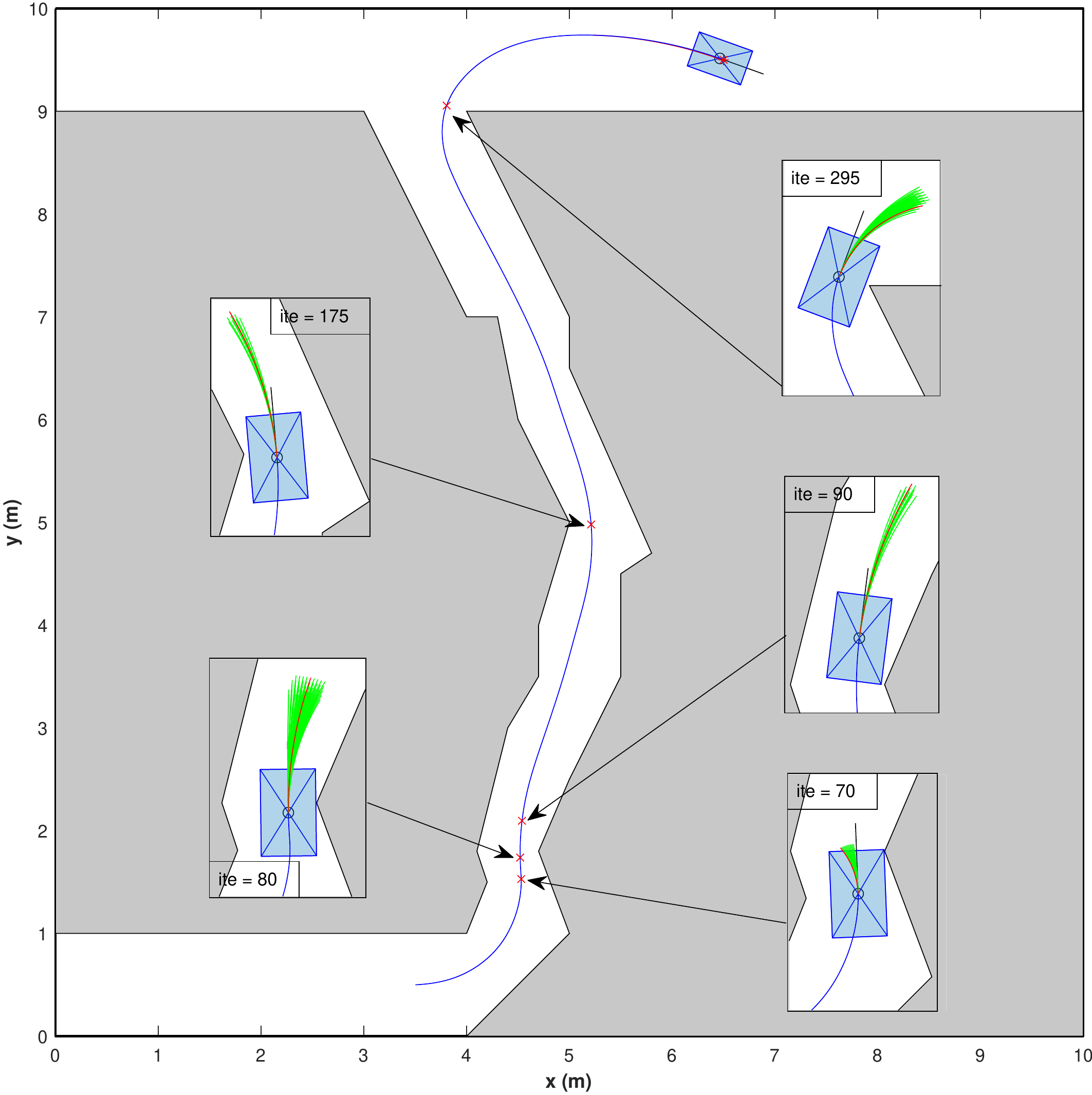}}
    \vspace{-5pt}
    \caption{The navigation system by DWA-CCD. The AGV is $0.55\times0.35m$ and the goal is located at $[6.5,9.5]$, depicted as the red star.}
    \vspace{-15pt}
    \label{fig:dwa-ccd}
\end{figure}

As depicted in \prettyref{fig:dwa-ccd}, we simulate a navigation system and test the AGV equipped with the 2D LiDAR in two environments. Following the pipeline of \prettyref{fig:workflow}, the perceived environment from the LiDAR is modeled as a myriad of edges. Furthermore, we combine our CCD framework with the local planner DWA\cite{dwa}, denoted by DWA-CCD, to navigate the AGV. It should be noted that trajectories generated by DWA are the set of arc segments due to dynamic constraints. So in this experiment, it will demonstrate that our CCD framework also works for non-polynomial trajectories in the use of Taylor series.

As shown in \prettyref{fig:sparseObs2}, the DWA-CCD will generate numerous candidate trajectories (in green) according to the robot's constraints and current status. Then it needs to find the optimal one (in red) based on several evaluation metrics, one of which is the collision information provided by our CCD framework. In DWA-CCD, we will continuously check the conflicts  for the AGV along candidate trajectories, without the risk of missing collisions resulting from the discretized interference detection method of the original DWA.
In addition, it is no need to shrink the robot model and inflate the environment in DWA-CCD, since our method describes the robots as full-dimensional objects, instead of the point-mass models in the original DWA. As an example in \prettyref{fig:narrowTube}, the dilated obstacles will totally block the corridor and therefore no path can be found by the DWA. However, benefiting from our CCD approach, the polygonal AGV can safely and easily pass through the extremely narrow corridor by the DWA-CCD.

\section{Conclusion and future work}
\label{sec:conclusion}
In this paper, we propose a generalized framework about CCD. We convert the collision scenarios between the robot and ellipsoid, sphere, cylinder and polyhedron into basic collision cases of edge-point, edge-edge and edge-triangle. Furthermore, we can transform these collision conditions into a set of polynomial inequalities, whose roots can provide the exact the collision instants. In our framework, we have tested three different mobile robots with various nonlinear kinematic and dynamic constraints in polynomial motions. Even for non-polynomial trajectories, our method also works within acceptable accuracy. 
Future work may include the abstraction method from the sensor data to the proposed obstacle models in real applications, and further combine our framework with the complicated planners.


\bibliographystyle{IEEEtran}
\typeout{}
\bibliography{mybibfiles}
\end{document}